%% file: main_final.tex
\documentclass[runningheads]{llncs}
\usepackage{graphicx}
\usepackage{tikz}
\usepackage{comment}
\usepackage{amsmath,amssymb} % define this before the line numbering.
\usepackage{color}

\usepackage[accsupp]{axessibility}  % Improves PDF readability for those with disabilities.

\usepackage{url}
\usepackage{stfloats}
\usepackage{subfigure}
\usepackage[colorlinks,linkcolor=red]{hyperref}
\usepackage{bm}
\usepackage{chngcntr}
\usepackage{multirow}

\newcommand\blfootnote[1]{%
  \begingroup
  \renewcommand\thefootnote{}\footnote{#1}%
  \addtocounter{footnote}{-1}%
  \endgroup
}

\begin{document}
% \renewcommand\thelinenumber{\color[rgb]{0.2,0.5,0.8}\normalfont\sffamily\scriptsize\arabic{linenumber}\color[rgb]{0,0,0}}
% \renewcommand\makeLineNumber {\hss\thelinenumber\ \hspace{6mm} \rlap{\hskip\textwidth\ \hspace{6.5mm}\thelinenumber}}
% \linenumbers

\newcommand{\he}[1]{{\color{blue}[He: #1]}}
\newcommand{\jiyao}[1]{{\color{blue}[Jiyao: #1]}}
\newcommand{\qiyu}[1]{{\color{blue}[Qiyu: #1]}}
\newcommand{\verify}[1]{{\color{red}[Verify: #1]}}
\newcommand{\todo}[1]{{\color{red}{\textbf{TODO}: #1}}}

\pagestyle{headings}
\mainmatter
\def\ECCVSubNumber{1831}  % Insert your submission number here

%\title{Bridging the Sim2Real Gap of RGBD-based Object Perception and Grasping by Depth Sensor Simulation and Domain Randomization} % Replace with your title

\title{Domain Randomization-Enhanced Depth Simulation and Restoration for
Perceiving and Grasping Specular and Transparent Objects}

% INITIAL SUBMISSION 
\begin{comment}
\titlerunning{ECCV-22 submission ID \ECCVSubNumber} 
\authorrunning{ECCV-22 submission ID \ECCVSubNumber} 
\author{Anonymous ECCV submission}
\institute{Paper ID \ECCVSubNumber}
\end{comment}
%******************

% CAMERA READY SUBMISSION
%\begin{comment}

\titlerunning{Domain Randomization-Enhanced Depth Simulation and Restoration}
% If the paper title is too long for the running head, you can set
% an abbreviated paper title here
%

\author{
Qiyu Dai\inst{1,*} \and
Jiyao Zhang\inst{2,*} \and
Qiwei Li\inst{1} \and
Tianhao Wu\inst{1} \and
Hao Dong\inst{1} \and \\
Ziyuan Liu\inst{3} \and
Ping Tan\inst{3,4} \and
He Wang\inst{1,\dagger}
}
%
%\authorrunning{Q. Dai, J. Zhang, et al.}
\authorrunning{Q. Dai, J. Zhang, Q. Li, T. Wu, H. Dong, Z. Liu, P. Tan, H. Wang}

\institute{ 
\mbox{Peking University \and Xi'an Jiaotong University}
\mbox{\and Alibaba XR Lab \and Simon Fraser University}
\email{\{qiyudai,lqw,hao.dong,hewang\}@pku.edu.cn},\\
\email{zhangjiyao@stu.xjtu.edu.cn}, 
\email{thwu@stu.pku.edu.cn}, \\
\email{ziyuan-liu@outlook.com}, 
\email{pingtan@sfu.ca}
}

%\end{comment}
%******************
\maketitle

%%%%%%%%% BODY TEXT
%\vspace{-12mm}
\input{tex_final/00_abstract.tex}

%\vspace{-10mm}
\section{Introduction}
%\vspace{-3mm}
\input{tex_final/01_intro.tex}

%\vspace{-5mm}
\section{Related Work}
%\vspace{-2mm}
\input{tex_final/02_related.tex}

%\vspace{-4mm}
\section{Domain Randomization-Enhanced Depth Simulation}
%\vspace{-2mm}
\input{tex_final/03_dreds}

%\vspace{-2mm}
\section{STD Dataset}
\input{tex_final/04_dts}

%\vspace{-3mm}
\section{Method}
\input{tex_final/05_method.tex}

\section{Tasks, Benchmarks and Results}
\input{tex_final/06_experiment.tex}
\section{Conclusions}
\input{tex_final/07_conclusion.tex}

%\clearpage\mbox{}Page \thepage\ of the manuscript.
%\clearpage\mbox{}Page \thepage\ of the manuscript.

%This is the last page of the manuscript.
%\par\vfill\par
%Now we have reached the maximum size of the ECCV 2022 submission (excluding references).
%References should start immediately after the main text, but can continue on p.15 if needed.

\clearpage
% ---- Bibliography ----
%
% BibTeX users should specify bibliography style 'splncs04'.
% References will then be sorted and formatted in the correct style.
%
\bibliographystyle{splncs04}
\bibliography{egbib}
%\end{document}

\clearpage

% \begin{document}
% \title{Supplementary Material for \\Domain Randomization-Enhanced Depth Simulation and Restoration for Perceiving and Grasping Specular and Transparent Objects}
% \author{}
% \institute{}
% \maketitle
{\renewcommand\baselinestretch{1.5}\selectfont
\begin{center}
\textbf{\Large Supplementary Material for \\Domain Randomization-Enhanced Depth Simulation and Restoration for Perceiving and Grasping Specular and Transparent Objects}
\end{center}
\par}

\input{supptex_final/08_supp_abstract}

\setcounter{section}{0}
\vspace{-3mm}
\section{Domain Randomization Details}
\vspace{-2mm}
\label{sec:secsupp1}
\input{supptex_final/09_supp_dr}

\vspace{-3mm}
\section{Network Implementation Details}
\vspace{-3mm}
\label{sec:secsupp3}
\input{supptex_final/11_supp_implement}

\vspace{-3mm}
\section{Additional Experiments and Results}
\label{sec:secsupp4}
\input{supptex_final/12_supp_experiment}

%\vspace{-3mm}

\vspace{-3mm}
\section{Additional Dataset Details}
\label{sec:secsupp2}
\input{supptex_final/10_supp_dataset}

\clearpage
% \bibliographystyle{splncs04}
% \bibliography{egbib}

\end{document}

%% file: tex_final/00_abstract.tex
\begin{abstract}
Commercial depth sensors usually generate noisy and missing depths, especially on specular and transparent objects,  which poses critical issues to downstream depth or point cloud-based tasks.
To mitigate this problem, we propose a powerful RGBD fusion network, SwinDRNet, for depth restoration.
We further propose Domain Randomization-Enhanced Depth Simulation (DREDS) approach to simulate an active stereo depth system using physically based rendering and generate a large-scale synthetic dataset that contains 130K photorealistic RGB images along with their simulated depths carrying realistic sensor noises. 
To evaluate depth restoration methods, we also curate a real-world dataset, namely STD, that captures 30 cluttered scenes composed of 50 objects with different materials from specular, transparent, to diffuse. 
Experiments demonstrate that the proposed DREDS dataset bridges the sim-to-real domain gap such that, trained on DREDS, our SwinDRNet can seamlessly generalize to other real depth datasets, e.g. ClearGrasp, and outperform the competing methods on depth restoration with a real-time speed.
We further show that our depth restoration effectively boosts the performance of downstream tasks, including category-level pose estimation and grasping tasks. Our data and code are available at \url{https://github.com/PKU-EPIC/DREDS}.

\keywords{Depth sensor simulation, specular and transparent objects,  domain randomization, pose estimation, grasping}
\blfootnote{*: equal contributions, $\dagger$: corresponding author}

\end{abstract}

%% file: tex_final/01_intro.tex
\begin{figure}[h]
\centering
\centering % trim = <left low right upper
	\includegraphics[trim=0 0 0 0,clip, width=\linewidth]
{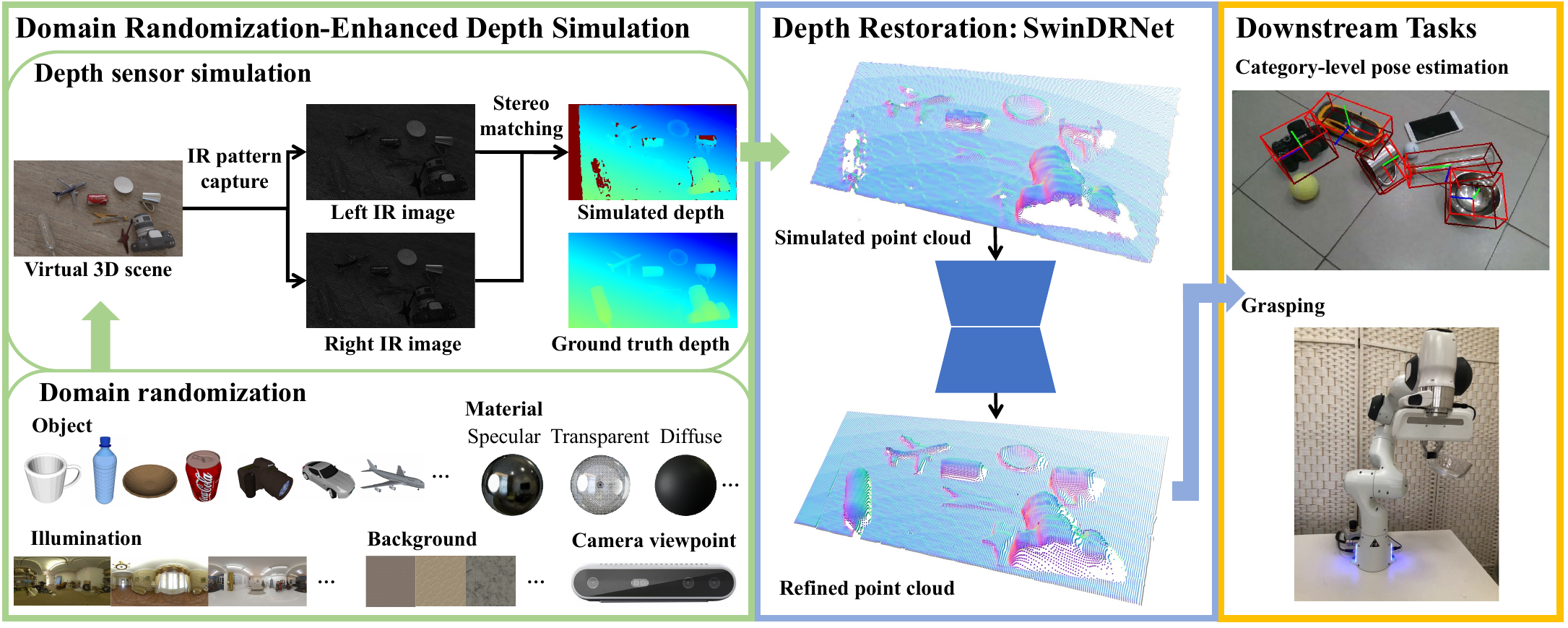}
%\vspace{-7mm}
\caption{\textbf{Framework overview.} From the left to right: we leverage domain randomization-enhanced depth simulation to generate paired data, on which we can train our depth restoration network SwinDRNet, and the restored depths will be fed to downstream tasks and improve estimating category-level pose and grasping for specular and transparent objects.}
\label{fig: teaser}
%\vspace{-7mm}
\end{figure}

With the emerging depth-sensing technologies, depth sensors and 3D point cloud data become more and more accessible, rendering many applications in VR/AR and robotics. 
Compared with RGB images, depth images or point clouds contain the true 3D information of the underlying scene geometry, thus depth cameras have been widely deployed in many robotic systems, \textit{e.g.} for object grasping~\cite{breyer2020volumetric,jiang2021synergies} and manipulation~\cite{weng2021captra,mu2021maniskill,mo2021where2act}, that care about the accurate scene geometry.
However, an apparent disadvantage of accessible depth cameras is that they may carry non-ignorable sensor noises more significant than usual noises in colored images captured by commercial RGB cameras. 
A more drastic failure case of depth sensing would be on objects that are either transparent or their surfaces are highly specular, where the captured depths would be highly erroneous and even missing around the specular or transparent region.
It should be noted that specular and transparent objects are indeed ubiquitous in our daily life, given most of the metallic surfaces are specular and many man-made objects are made of glasses and plastics which can be transparent.
The existence of so many specular and transparent objects in our real-world scenes thus poses severe challenges to depth-based vision systems and limits their application scenarios to well-controlled scenes and objects made of diffuse materials.

In this work, we devise a two-stream Swin Transformer~\cite{liu2021swin} based RGB-D fusion network, SwinDRNet, for learning to perform depth restoration. 
However, it is a lack of real data composed of paired sensor depths and perfect depths to train such a network. 
Previous works on depth completion for transparent objects, like ClearGrasp~\cite{sajjan2020clear} and LIDF~\cite{zhu2021rgb}, leverage synthetic perfect depth image for network training. They simply remove the transparent area in the perfect depth and their methods then learn to complete the missing depths in a feedforward way or further combines with depth optimization.
We argue that both the methods can only access incomplete depth images during training and never see a depth with realistic sensor noises, leading to suboptimality when directly deployed on real sensor depths. 
Also, these two works only consider a small number of similar objects with little shape variations and all being transparent and hence fail to demonstrate their usefulness when adopted in scenes with completely novel object instances.
Given material specularity or transparency forms a continuous spectrum, it is further questionable whether their methods can handle objects of intermediate transparency or specularity.

To mitigate the problems in the existing works, we thus propose to synthesize depths with realistic sensor noise patterns by simulating an active stereo depth camera resembling RealSense D415. Our simulator is built on Blender and leverages raytracing to mimic the IR stereo patterns and compute the depths from them. To facilitate generalization, we further adopt domain randomization techniques that randomize the object textures, object materials (from specular, transparent, to diffuse), object layout, floor textures, illuminations along camera poses. This domain randomization-enhanced depth simulation method, or in short DREDS, leads to 130K photorealistic RGB images and their corresponding simulated depths. We further curate a real-world dataset, STD dataset, that contains 50 objects with specular, transparent, and diffuse material. Our extensive experiments demonstrate that our SwinDRNet trained on DREDS dataset can handle depth restoration on object instances from both seen and unseen object categories in STD dataset and can even seamlessly generalize to ClearGrasp dataset and beat the previous state-of-the-art method, LIDF~\cite{zhu2021rgb} trained on ClearGrasp dataset. Additionally, SwinDRNet allows real-time depth restoration (30 FPS). Our further experiments on estimating category-level pose and grasping specular and transparent objects prove that our depth restoration is both generalizable and successful.

%% file: tex_final/02_related.tex
\subsection{Depth Estimation and Restoration}
%\vspace{-2mm}
The increasing popularity of RGBD sensors has encouraged much research on depth estimation and restoration. 
% RGB-only
Many works~\cite{eigen2014depth,jiao2018look,long2021adaptive} directly estimate the depth from a monocular RGB image, but fail to restore accurate geometries of the point cloud because of the few geometric constraints of the color image.
% RGB, sparse depth
Other studies~\cite{park2020non,xiong2020sparse,qu2021bayesian} restore the dense depth map given the RGB image and the sparse depth from LiDAR, but the estimated depth still suffers from low quality due to the limited geometric guidance of the sparse input.
% RGB, noisy and imcomplete depth
Recent research focuses on commercial depth sensors, trying to complete and refine the depth values from the RGB and noisy dense depth images.
Sajjan \emph{et al.}~\cite{sajjan2020clear} proposed a two-stage method for transparent object depth restoration, which firstly estimates surface normals, occlusion boundaries, and segmentations from RGB images, and then calculates the refined depths via global optimization. However, the optimization is time-consuming, and heavily relies on the previous network predictions.
Zhu \emph{et al.}~\cite{zhu2021rgb} proposed an implicit transparent object depth completion model, including the implicit representation learning from ray-voxel pairs and the self-iterating refinement, but voxelization of the 3D space results in heavy geometric discontinuity of the refined point cloud.
Our method falls into this category and outperforms those methods, ensuring fast inference time and better geometries to improve the performance of downstream tasks.

%\vspace{-4mm}
\subsection{Depth Sensor Simulation}
%\vspace{-2mm}
To close the sim-to-real gap, the recent research focuses on generating simulated depth maps with realistic noise distribution.~\cite{landau2015simulating} simulated the pattern projection and capture system of Kinect to obtain simulated IR images and perform stereo matching, but could not simulate the sensor noise caused by object materials and scene environments.~\cite{planche2017depthsynth} proposed an end-to-end framework to simulate the mechanism of various types of depth sensors. However, the rasterization method limits the
photorealistic rendering and physically correct simulation.~\cite{planche2021physics} presented a new differentiable structure-light depth sensor simulation pipeline, but cannot simulate the transparent material, limited by the renderer.
Recently,~\cite{zhang2022close} proposed a physics-grounded active stereovision depth sensor simulator for various sim-to-real applications, but focused on instance-level objects and the robot arm workspace.
Our DREDS pipeline generates realistic RGBD images for various materials and scene environments, which can generalize the proposed model to category-level unseen object instances and novel categories.

%\vspace{-4mm}
\subsection{Domain Randomization}
%\vspace{-2mm}
Domain randomization bridges the sim-to-real gap in the way of data augmentation. Tobin \emph{et al.}~\cite{tobin2017domain} first explore transferring to real environments by generating training data through domain randomization. 
Subsequent works~\cite{tremblay2018training,yue2019domain,prakash2019structured} generate synthetic data with sufficient variation by manually setting randomized features. Other studies~\cite{zakharov2019deceptionnet} perform randomization using the neural networks. These works have verified the effectiveness of domain randomization on the tasks such as robotic manipulation~\cite{peng2018sim}, object detection and pose estimation~\cite{khirodkar2019domain}, \emph{etc}. In this work, we combine the depth sensor simulation pipeline with domain randomization, which, for the first time, enables direct generalization to unseen diverse real instances on specular and transparent object depth restoration.
%efficiently enhances the generalization ability to the real-world environment.

%% file: tex_final/03_dreds.tex
\subsection{Overview}
%\vspace{-1mm}
In this work, we propose a simulated RGBD data generation pipeline, namely Domain Randomization Enhanced Depth Simulation (DREDS), for tasks of depth restoration, object perception, and robotic grasping. We build a depth sensor simulator, modeling the mechanism of the active stereo vision depth camera system based on the physically based rendering, along with the domain randomization technique to handle real-world variations. 

\begin{table}
\scriptsize
\renewcommand{\arraystretch}{1.2} % Default value: 1
\begin{center}
\caption{\textbf{Comparisons of specular and transparent depth restoration dataset.} S, T, and D refer to specular, transparent, and diffuse materials, respectively. \#Objects refers to the number of objects. SN+CG means the objects are selected from ShapeNet and ClearGrasp (the number are not mentioned).}
\label{table:depth_restoration_dataset}
% \begin{tabular}{c|c|c|c|c|c|c|c|c|c|c|c}
\begin{tabular}{c|c|c|c|c}
\hline
% Dataset & Type  & \multicolumn{7}{|c|}{Data \& Annotations}                    & \#Obj & Material& Size\\ \hline
%         &       & RGB & IR & Depth & GT Depth & Ins Mask & Normal & NOCS map &       &         &\\ \hline
% CG-Syn~\cite{sajjan2020clear}   & Syn & \checkmark  &            & Syn & \checkmark &            & \checkmark &            & 9 & T & 50K \\ 
% Omniverse~\cite{zhu2021rgb}     & Syn & \checkmark  &            & Syn & \checkmark &            &            &            & SN+CG & T+D & 60K \\ 
% CG-Real~\cite{sajjan2020clear}  & Real & \checkmark &            & Real& \checkmark &            & \checkmark &            & 10 & T & 286 \\ 
% TODD~\cite{xu2021seeing}        & Real & \checkmark &            & Real& \checkmark & \checkmark &            &            & 6 & T & 1.5K \\ \hline
% \textbf{DREDS}                  & Sim & \checkmark  & \checkmark & Sim & \checkmark & \checkmark & \checkmark & \checkmark & 1,861 & S+T+D & 130K \\ 
% \textbf{STD}                    & Real & \checkmark &            & Real& \checkmark & \checkmark &            & \checkmark & 50 & S+T+D & 27K \\ \hline
Dataset & Type & \#Objects & Type of Material & Size\\ \hline
ClearGrasp-Syn~\cite{sajjan2020clear} & Syn & 9 & T & 50K \\ 
Omniverse~\cite{zhu2021rgb} & Syn & SN+CG & T+D & 60K \\ 
ClearGrasp-Real~\cite{sajjan2020clear} & Real & 10 & T & 286 \\ 
TODD~\cite{xu2021seeing} & Real & 6 & T & 1.5K \\ \hline
\textbf{DREDS} & Sim & 1,861 & S+T+D & 130K \\ 
\textbf{STD} & Real & 50 & S+T+D & 27K \\ \hline
\end{tabular}
\end{center}
%\vspace{-11mm}
\end{table}

Leveraging domain randomization and active stereo sensor simulation, we present DREDS, the large-scale simulated RGBD dataset, containing photorealistic RGB images and depth maps with the real-world measurement noise and error, especially for the hand-scale objects with specular and transparent materials. 
The proposed DREDS dataset bridges the sim-to-real domain gap, and generalizes the RGBD algorithms to unseen objects. DREDS dataset's comparison to the existing specular and transparent depth restoration datasets is summarized in Table \ref{table:depth_restoration_dataset}.

%\vspace{-4mm}
\subsection{Depth Sensor Simulation}
%\vspace{-1mm}
% real sensor mechanisms
A classical active stereo depth camera system contains an infrared (IR) projector, left and right IR stereo cameras, and a color camera. To measure the depth, the projector emits an IR pattern with dense dots to the scene. Subsequently, the two stereo cameras capture the left and right IR images, respectively. Finally, the stereo matching algorithm is used to calculate per-pixel depth values based on the discrepancy between the stereo images, to get the final depth scan.
Our depth sensor simulator follows this mechanism, containing light pattern projection, capture, and stereo matching. The simulator is mainly built upon Blender~\cite{Blender}. 

\textbf{Light Pattern Capture via physically based rendering.}
For real-world specular and transparent objects, the IR light from the projector may not be received by the stereo cameras, due to the reflection on the surface or the refraction through the transparent objects, resulting in inaccurate and missing depths. 
To simulate the physically correct IR pattern emission and capture process, we thus adopt physically based ray tracing, a technique that mimics the real light transportation process, and supports various surface materials especially specular and transparent materials. 

Specifically, the textured spotlight projects a binary pattern image into the virtual scene. Sequentially, the binocular IR images are rendered from the stereo cameras. We manage to simulate IR images via visible light rendering, where both the light pattern and the reduced environment illumination contribute to the IR rendering. From the perspective of physics, the difference between IR and visible light is the reflectivity and refractive index of the object. We note that the wavelength (850 nm) of IR light used in depth sensors, \textit{e.g.} RealSense D415, is close to the visible light (400-800 nm). So the resulting effects have already been well-covered by the randomization in object reflectivity and refractive index used in DREDS, which constructs a superset of real IR images. To mimic the portion of IR in environmental light, we reduce its intensity. Finally, all RGB values are converted to intensity, which is our final IR image.

\textbf{Stereo Matching.}
We perform stereo matching to obtain the disparity map, which can be transferred to the depth map leveraging the intrinsic parameters of the depth sensor. In detail, we compute a matching cost volume over the left and right IR images along the epipolar line and find the matching results with minimum matching cost. Then we perform sub-pixel detection to generate a more accurate disparity map using the quadratic curve fitting method. To generate a more realistic depth map, we perform post-processing, including left/right consistency check, uniqueness constraint, median filtering, \emph{etc}.

%\vspace{-2mm}
\subsection{Simulated Data Generation with Domain Randomization}
Based on the proposed depth sensor simulator, we formulate the simulated RGBD data generation pipeline as 
%\begin{equation} 
$D = Sim(\mathcal{S}, \mathcal{C})$, 
%\end{equation} 
where $\mathcal{S} = \{\mathcal{O}, \mathcal{M}, \mathcal{L}, \mathcal{B}\}$ denotes scene-related simulation parameters in the virtual environment, including $\mathcal{O}$ the setting of the objects with random categories, poses, arrangements, and scales, $\mathcal{M}$ the setting of random object materials from specular, transparent, to diffuse, $\mathcal{L}$ the setting of environment lighting from varying scenes with different intensities, $\mathcal{B}$ the setting of background floor with diverse materials. $\mathcal{C}$ is the cameras' statue parameters, consisting of intrinsic and extrinsic parameters, the pattern image, baseline distance, \emph{etc}. Taking these settings as input, the proposed simulator $Sim$ generates the realistic RGB and depth images $D$.

To construct scenes with sufficient variations so that the proposed method can generalize to the real, we adopt domain randomization to enhance the generation, considering all these aspects. See supplementary materials for more details.

%\vspace{-3mm}
\subsection{Simulated Dataset: DREDS}
Making use of domain randomization and depth simulation, we construct the large-scale simulated dataset, DREDS.
In total, DREDS dataset consists of two subsets: 1) \textbf{DREDS-CatKnown}: 100,200 training and 19,380 testing RGBD images made of 1,801 objects spanning 7 categories from ShapeNetCore~\cite{chang2015shapenet}, with randomized specular, transparent, and diffuse materials, 2) \textbf{DREDS-CatNovel}: 11,520 images of 60 category-novel objects, which is transformed from GraspNet-1Billion~\cite{fang2020graspnet} that contains CAD models and annotates poses, by changing their object materials to specular or transparent, to verify the ability of our method to generalize to new object categories. Examples of paired simulated RGBD images of DREDS-Catknown and DREDS-CatNovel datasets are shown in Figure~\ref{fig:simexam}.

%% file: tex_final/04_dts.tex
\begin{figure}[htbp]
%\centering
\begin{minipage}[t]{0.45\textwidth}
%\centering
\includegraphics[width=4.5cm]{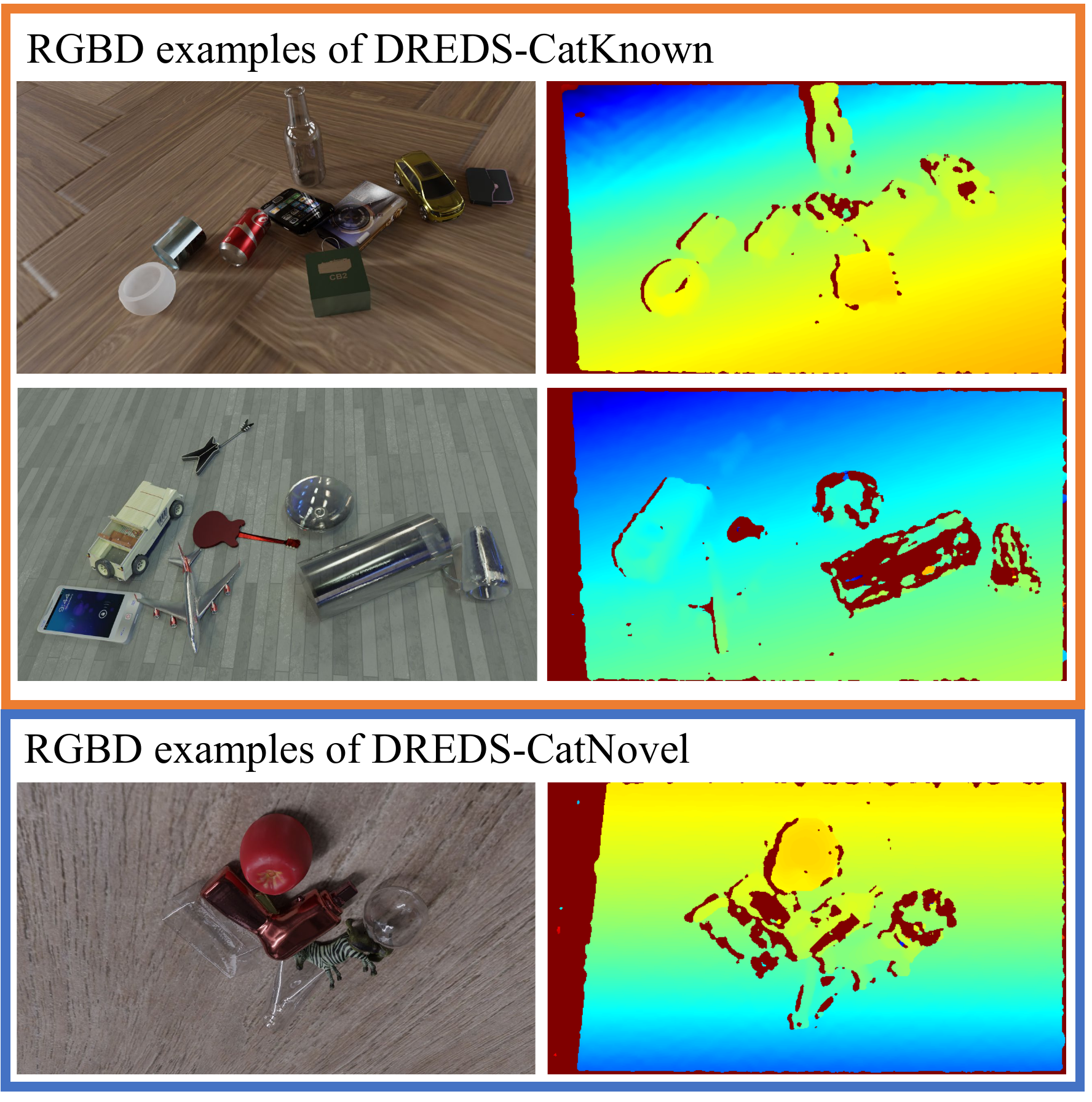}
%%\vspace{-5mm}
\caption{\textbf{RGBD examples of \\DREDS dataset.}}
\label{fig:simexam}
\end{minipage}
%\hspace{4mm}
\begin{minipage}[t]{0.54\textwidth}
\centering
\includegraphics[width=6.65cm]{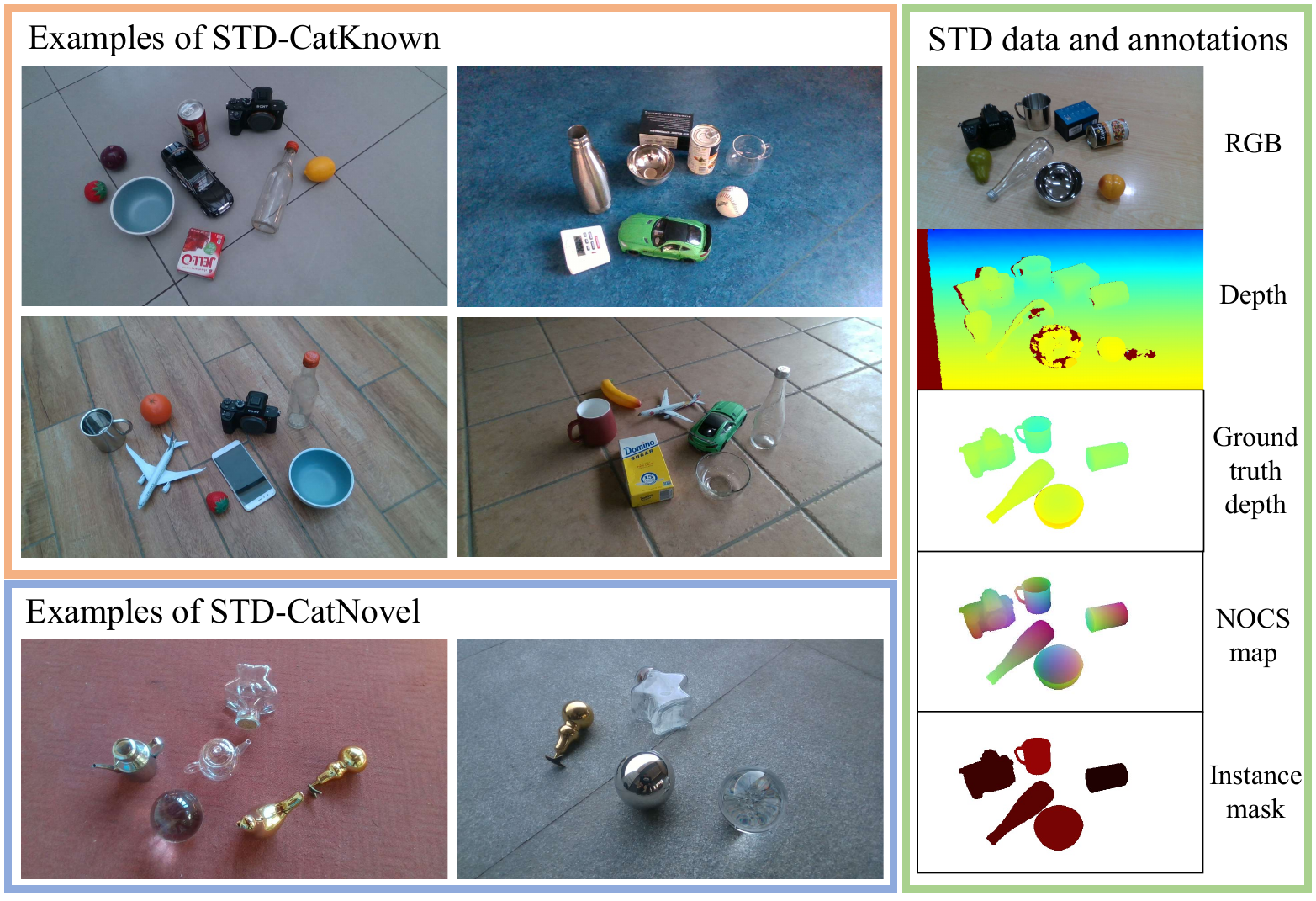}
%%\vspace{-5mm}
\caption{\textbf{Scene examples and annotations of STD dataset.}}
\label{fig:realexam}
\end{minipage}
%%\vspace{-4mm}
\end{figure}

% %\vspace{-6mm}
\subsection{Real-world Dataset: STD}
% %\vspace{-6mm}
To further examine the proposed method in real scenes, we curate a real-world dataset, composed of Specular, Transparent, and Diffuse objects, which we call it STD dataset. Similar to DREDS dataset, STD dataset contains 1) \textbf{STD-CatKnown}: the subset with category-level objects, for the evaluation of depth restoration and category-level pose estimation tasks, and 2) \textbf{STD-CatNovel}: the subset with category-novel objects for evaluating the generalization ability of the proposed SwinDRNet method. 
Figure~\ref{fig:realexam} shows the scene examples and annotations of STD dataset.

%\vspace{-2mm}
\subsection{Data Collection}
% %\vspace{-6mm}
We collect an object set, covering specular, transparent, and diffuse materials. Specifically, for STD-CatKnown dataset, we collect 42 instances from 7 known ShapeNetCore~\cite{chang2015shapenet} categories, and several category-unseen objects from the YCB dataset~\cite{calli2017yale} and our own as the distractors. For STD-CatNovel dataset, we pick 8 specular and transparent objects from unseen categories.
For each object except the distractors, we utilize the photogrammetry-based reconstruction tool, Object Capture API~\cite{macOSAPI}, to obtain its clean and accurate 3D mesh for ground truth poses annotation, so that we can yield ground truth depth and object masks.

We capture data from 30 different scenes (25 for STD-CatKnown, 5 for STD-CatNovel) with various backgrounds and illuminations, using RealSense D415. In each scene, over 4 objects with random arrangements are placed in a cluttered way. The sensor moves around the objects in an arbitrary trajectory.
In total, we take 22,500 RGBD frames for STD-CatKnown, and 4,500 for STD-CatNovel. 

Overall, the proposed real-world STD dataset consists of 27K RGBD frames, 30 diverse scenes, and 50 category-level and category-novel objects, making it facilitate the further generalizable object perception and grasping research.

%% file: tex_final/05_method.tex
In this section, we introduce our network for depth restoration in section \ref{sec:depth_restore} and then introduce the methods we used for downstream tasks, \textit{i.e.} category-level 6D object pose estimation and robotic grasping, in section \ref{sec:downstream}.

\begin{figure}
\centering
\includegraphics[scale=0.38]{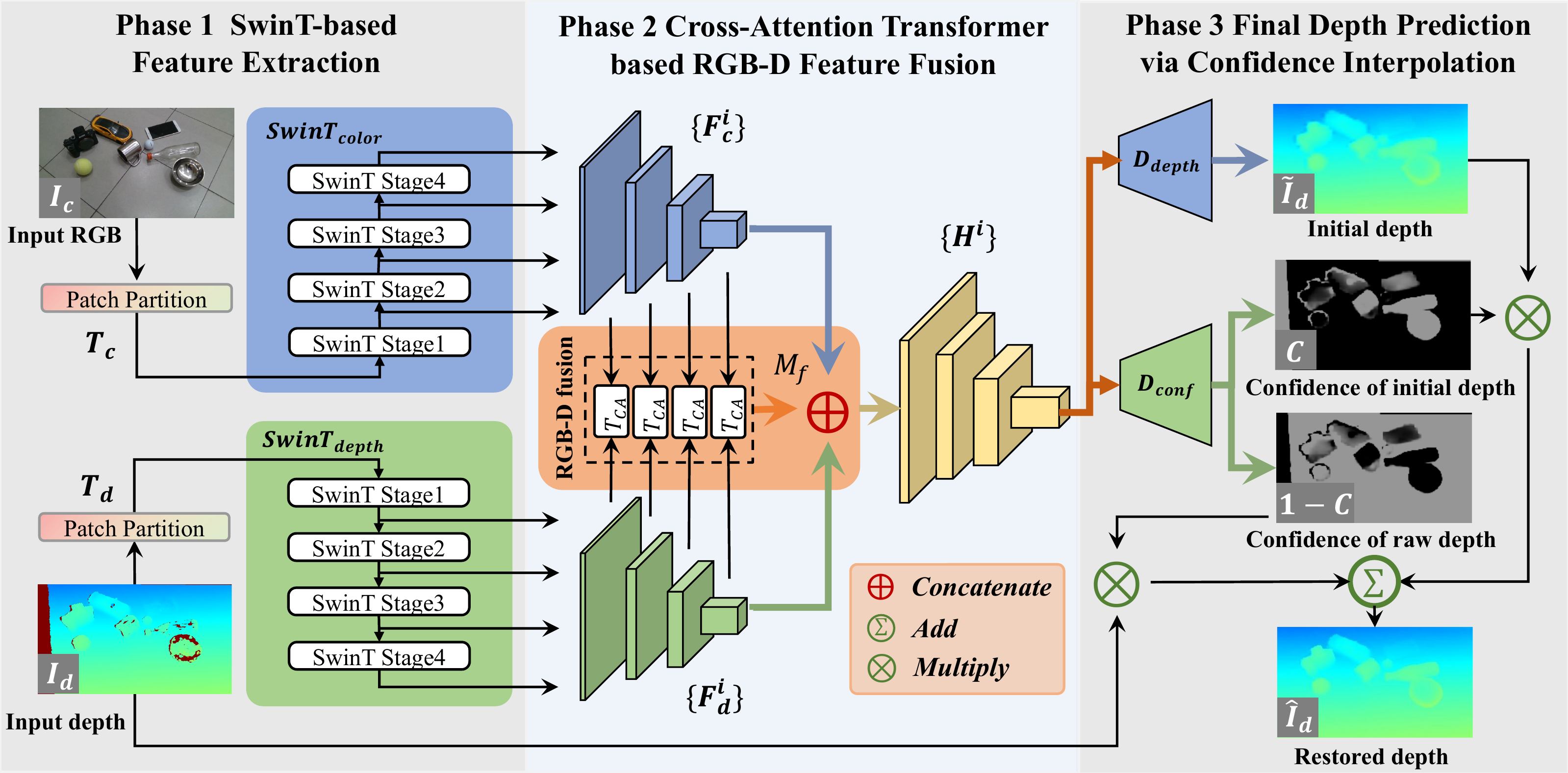}
\caption{\textbf{Overview of our proposed depth restoration network SwinDRNet.} We first extract the multi-scale features of RGB and depth image in phase 1, respectively. Next, in phase 2, our network fuse features of different modalities. Finally, we generate the initial depth map and confidence maps via two decoders, respectively, and fuse the raw depth and initial depth using the predicted confidence map.}
\label{fig: overview of SwinDR}
\end{figure}

\subsection{SwinDRNet for Depth Restoration}
\label{sec:depth_restore}
\textbf{Overview.} To restore the noisy and incomplete depth, we propose a SwinTransformer~\cite{liu2021swin} based depth restoration network, namely \textbf{SwinDRNet}.
%We devise an end-to-end learning framework for depth restoration, namely \emph{SwinT-based\cite{liu2021swin} Depth Restoration Net} (\textbf{SwinDR}).

SwinDRNet takes as input a RGB image $\ \mathcal{I}_{c} \in \mathbb{R}^{H\times W\times 3}$ along with its aligned depth image $\mathcal{I}_{d} \in \mathbb{R}^{H\times W}$ and outputs a refined depth $\hat{ {\mathcal{I}_d}} \in \mathbb{R}^{H\times W}$ that restores the error area of the depth image and completes the invalid area, where $H$ and $W$ are the input image sizes.

We notice that prior works, \emph{e.g.} PVN3D~\cite{he2020pvn3d}, usually leverage a heterogeneous architecture that extracts CNN features from RGB and extracts PointNet++~\cite{qi2017pointnet++} features from depth. We, for the first time, devise a homogeneous and mirrored architecture that only leverages SwinT to extract and hierarchically fuse the RGB and depth features.

As shown in Figure \ref{fig: overview of SwinDR}, the architecture of SwinDRNet is a two-stream fused encoder-decoder and can be further divided into three phases:
in the first phase of feature extraction, we leverage two separate SwinT backbones to extract hierarchical features $\{\mathcal{F}_{c}^i\}$ and $\{\mathcal{F}_{d}^i\}$ from the input RGB image $\mathcal{I}_{c}$ and depth $\mathcal{I}_{d}$, respectively;
In the second stage of RGBD feature fusion, we propose a fusion module $M_{f}$ that utilizes cross-attention transformers to combine the features from the two streams and generate 
fused hierarchical features $\{\mathcal{H}^i\}$ ;
and finally in the third phase, we propose two decoder modules,
the depth decoder module $D_{depth}$ decodes the fused feature into a raw depth and the confidence decoder module $D_{conf}$ outputs a confidence map of the predicted raw depth, and from the outputs we can compute the final restored depth by using the confidence map to select accurate depth predictions at noisy and invalid areas of the input depth while keeping the originally correct area as much as possible.

\textbf{SwinT-based Feature Extraction.} To accurately restore the noisy and incomplete depth, we need to leverage visual cues from the RGB image that helps depth completion as well as geometric cues from the depth that may save efforts at areas with correct input depths. 
To extract rich features, we propose to utilize SwinT~\cite{liu2021swin} as our backbone, since it is a very powerful and efficient network that can produce hierarchical feature representations at different resolutions and has linear computational complexity with respect to input image size.
Given our inputs contain two modalities -- RGB and depth, 
we deploy two seperate SwinT networks, $SwinT_\text{color}$ and $SwinT_\text{depth}$, to extract features from $\mathcal{I}_c$ and $\mathcal{I}_d$, respectively.
For each one of them, we basically follow the design of SwinT. 
Taking the $SwinT_\text{color}$ as an example: we first divide the input RGB image $\ \mathcal{I}_{c} \in \mathbb{R}^{H\times W\times 3}$ into non-overlapping patches, which is also called tokens, $\mathcal{T}_{c} \in \mathbb{R}^{{\frac{H}{4}}\times {\frac{W}{4}}\times 48}$; we then pass $ {\mathcal{T}_{c}} $ through the four stages of SwinT to generate the multi-scale features $\{\mathcal{F}_{c}^i\}$, which are especially useful for dense depth prediction thanks to the hierarchical structure. The encoder process can be formulated as: 
\begin{equation} \{\mathcal{F}_c^i\}_{i=1,2,3,4} = SwinT_\text{color}(\mathcal{T}_{c}), \end{equation}
\begin{equation} \{\mathcal{F}_d^i\}_{i=1,2,3,4} = SwinT_\text{depth}(\mathcal{T}_d). \end{equation} 
%$\mathcal{T}_{r}$ and$\mathcal{T}_{d}$ denote the input tokens of RGB branch and depth branch, respectively. 
where %${\mathcal{F}_{r}} \in {\{ \mathcal{F}^i_r \}}_{i=1}^4$ is the generated multi-scale feature of RGB branch and 
$ {\mathcal{F}^i} \in \mathbb{R}^{{\frac{H}{4i}}\times {\frac{W}{4i}}\times iC} $ and $C$ is the output feature dimension of the linear embedding layer in the first stage of SwinT.

\textbf{Cross-Attention Transformer based RGB-D Feature Fusion.} Given the hierarchical features $\{\mathcal{F}_{c}^i\}$ and $\{\mathcal{F}_{d}^i\}$ from the two-stream SwinT backbone, our RGB-D fusion module $M_f$ leverages cross-attention transformers to fuse the corresponding $\mathcal{F}_{c}^i$ and  $\mathcal{F}_{d}^i$ into $\mathcal{H}^i$.
For attending feature $\mathcal{F_A}$ to $\mathcal{F_B}$, a common cross-attention transformer $T_{CA}$ first calculates the query vector $Q_A$ from $\mathcal{F}_A$ and the key $K_B$ and value $V_B$ vectors from feature $\mathcal{F}_B$:
\begin{equation} \ Q_A = \mathcal{F}_A \cdot W_q,  ~~K_B = \mathcal{F}_B \cdot W_k,  ~~V_B = \mathcal{F}_B \cdot W_v, \end{equation}
where $W$s are the learnable parameters,
and then computes the cross-attention feature $\mathcal{H}_{\mathcal{F}_A\rightarrow \mathcal{F}_B}$ from $\mathcal{F}_A$ to $\mathcal{F}_B$:
\begin{equation} \ \mathcal{H}_{\mathcal{F}_A\rightarrow \mathcal{F}_B} = T_{CA}(\mathcal{F}_A, \mathcal{F}_B)
=\text{softmax}\left(\frac{Q_A \cdot K_B^T}{\sqrt{d_K}}\right) \cdot V_B, \end{equation}
where $d_K$ is the dimension of $Q$ and $K$. 

In our module $M_f$, we leverage bidirectional cross-attention by deploying two cross-attention transformers to obtained the cross-attention features from both directions, and then concatenates them with the original features to form the fused hierarchical features $\{\mathcal{H}^i\}$, as shown below:
\begin{equation} \ \mathcal{H}^i = 
\mathcal{H}_{\mathcal{F}_c^i\rightarrow \mathcal{F}_d^i}
 \bigoplus \mathcal{H}_{\mathcal{F}_d^i\rightarrow \mathcal{F}_c^i} \bigoplus \mathcal{F}_c^i \bigoplus \mathcal{F}_d^i, \end{equation}
where $\bigoplus$ represents concatenation along the channel axis.

% \subsubsection{Depth Decoder $D_\text{depth}$ and Confidence Decoder $D_\text{conf}$.} % \todo{You need a few words about the two module as well as raw depth prediction 
% ${I}^{raw}_{d}$ and confidence map $C$}. 

\textbf{Final Depth Prediction via Confidence Interpolation.} The credible area of the input depth map (\emph{e.g.}, the edges of specular or transparent objects in contact with background or diffusive objects) plays a critical role in providing information about spatial arrangement. Inspired by the previous works~\cite{van2019sparse,hu2021penet}, we make use of a confidence map between the raw and predicted depth maps. However, unlike~\cite{van2019sparse,hu2021penet} predicting the confidence map between the multi-modality, we focus on preserving the correct original value to generate more realistic depth maps with less distortion. The final depth map can be formulated as:
\begin{equation} \hat{\mathcal{I}}_d = C  \bigotimes \tilde{\mathcal{I}}_{d} + (1-C) \bigotimes \mathcal{I}_{d} \end{equation} where $\bigotimes$ represents elementwise multiplication, and $\hat{\mathcal{I}}_d$ and $\tilde{\mathcal{I}}_{d}$ denote the final restored depth and the output of depth decoder head, respectively. 

\textbf{Loss Functions} 
For SwinDRNet training, we supervise both the final restored depth $\hat{\mathcal{I}}_d$ and the output of depth decoder head $\tilde{\mathcal{I}}_{d}$, which is formulated as:
\begin{equation}
    \mathcal{L} = \omega_{\tilde{\mathcal{I}}_d}\mathcal{L}_{\tilde{\mathcal{I}}_d} + \omega_{\hat{\mathcal{I}}_d}\mathcal{L}_{\hat{\mathcal{I}}_d},
\end{equation}
where $\mathcal{L}_{\hat{\mathcal{I}}_d}$ and $\mathcal{L}_{\tilde{\mathcal{I}}_d}$ are the losses of $\hat{\mathcal{I}}_d$ and $\tilde{\mathcal{I}}_{d}$, respectively. $\omega_{\hat{\mathcal{I}}_d}$ and $\omega_{\tilde{\mathcal{I}}_d}$ are weighting factors. Each of the two loss can be formulated as:
\begin{equation}
    \mathcal{L}_i = \omega_{n}\mathcal{L}_n + \omega_{d}\mathcal{L}_d + \omega_{g}\mathcal{L}_g,
\end{equation} 
where $\mathcal{L}_n$, $\mathcal{L}_d$ and $\mathcal{L}_g$ are the L1 losses between the predicted and ground truth surface normal, depth and the gradient map of depth image, respectively. $\omega_{n}$, $\omega_{d}$ and $\omega_{g}$ are the weights for different losses. We further add higher weight to the loss within the foreground region, to push the network to concentrate more on the objects. 

%\vspace{-2mm}
\subsection{Downstream Tasks}
\label{sec:downstream}
\textbf{Category-level 6D Object Pose Estimation.}
Inspired by~\cite{wang2019normalized}, we use the same backbone with SwinDRNet, and add two decoder heads to predict coordinates of the NOCS map and semantic segmentation mask. Then we follow the method~\cite{wang2019normalized}, perform pose fitting between the restored object point clouds in the world coordinate space and the predicted object point clouds in the normalized object coordinate space, and perform pose fitting to get the 6D object pose. 

\textbf{Robotic Grasping.} By combining SwinDRNet to the object grasping task, we can analyze the performance of depth restoration on the robotic manipulation. We adopt the end-to-end network, GraspNet-baseline~\cite{fang2020graspnet}, to predict the 6-DoF grasping poses directly from the scene point cloud. Given the restored depth map from SwinDRNet, the scene point cloud is transformed and sent to GraspNet-baseline. Then the model predicts the grasp candidates. Finally, the gripper of the parallel-jaw robot arm executes the target rotation and position selected from those candidates.

%% file: tex_final/06_experiment.tex
In this section, we train our SwinDRNet on the train split of DREDS-CatKnown dataset and deploy it on the tasks including category-level 6D object pose estimation and robotic grasping.

%\vspace{-5mm}
\subsection{Depth Restoration}
%\vspace{-3mm}

\textbf{Evaluation Metrics.} 
We follow the metrics of transparent objects depth completion in~\cite{zhu2021rgb}: 1) \textbf{RMSE}: the root mean squared error, 2) \textbf{REL}: the mean absolute relative difference, 3) \textbf{MAE}: the mean absolute error, 4) the percentage of $d_i$ satisfying $max(\frac{d_i}{d_i^*}, \frac{d_i^*}{d_i}) < \bm{\delta} $, where $d_i$ denotes the predicted depth, $d_i^*$ is GT and $\delta \in {\{1.05, 1.10, 1.25\}}$. We resize the prediction and GT to $126 \times 224$ resolution for fair comparisons, and evaluate in all objects area and challenging area (specular and transparent objects), respectively.

% \textbf{Comparison to State-of-the-Art Methods}
\textbf{Baselines.}
We compare our method with several state-of-the-art methods, including LIDF \cite{zhu2021rgb}, the SOTA method for depth completion of transparent objects, and NLSPN~\cite{park2020non}, the SOTA method for depth completion on NYUv2~\cite{uhrig2017sparsity} dataset. All baselines are trained on the train split of DREDS-CatKnown and evaluated on four types of testing data: 1) the test split of DREDS-CatKnown: simulated images of category-known objects. 2) DREDS-CatNovel: simulated images of category-novel objects. 3) STD-CatKnown: real images of category-known objects; 4) STD-CatNovel. real images of category-novel objects. 

\textbf{Results.} The quantitative results reported in Table \ref{table:depth_restoration_ourdataset} show that we achieve the best performance compared to other methods on DREDS and STD datasets, and have a powerful generalization ability to transfer to not only novel category objects in the simulation environment but also in the real world. 
In addition to performance gain, ours (30 FPS) is significantly faster than LIDF (13 FPS) and the two-branch baseline that uses PointNet++ on depth (6 FPS). Although it is a little slower than NLSPN (35 FPS), SwinDRNet has achieved real-time depth restoration, and our code still has room for optimization and speedup. The methods are all evaluated on an NVIDIA RTX 3090 GPU.

\begin{table}
%\vspace{-8mm}
\scriptsize
\renewcommand{\arraystretch}{1.2} % Default value: 1
\begin{center}
\caption{\textbf{Quantitative comparison to state-of-the-art methods on DREDS and STD.} $\downarrow$ means lower is better, $\uparrow$ means higher is better. The left of '/' shows the results evaluated on all objects, and the right of '/' shows the results evaluated on specular and transparent objects. Note that only one result is reported on STD-CatNovel, because all the objects are specular or transparent.}
%%\vspace{-7mm}
\label{table:depth_restoration_ourdataset}
\begin{tabular}{c|cccccc}
\hline%\noalign{\smallskip}
Methods & RMSE$\downarrow$ & REL$\downarrow$ & MAE$\downarrow$ & $\delta_{1.05}\uparrow$ & $\delta_{1.10}\uparrow$ & $\delta_{1.25}\uparrow$ \\
%\noalign{\smallskip}
\hline
%\noalign{\smallskip}
& \multicolumn{6}{c}{DREDS-CatKnown (Sim)} \\ \hline
NLSPN &\textbf{0.010}/0.011 & 0.009/0.011 &0.006/0.007 &97.48/96.41 &99.51/99.12 &99.97/99.74\\ \hline
LIDF &0.016/0.015 &0.018/0.017 &0.011/0.011 &93.60/94.45 &98.71/98.79 &99.92/99.90\\ \hline
Ours &\textbf{0.010/0.010} &\textbf{0.008/0.009} &\textbf{0.005/0.006} &\textbf{98.04/97.76} &\textbf{99.62/99.57} &\textbf{99.98/99.97}\\ \hline

& \multicolumn{6}{c}{DREDS-CatNovel (Sim)} \\ \hline
NLSPN &0.026/0.031 &0.039/0.054 &0.015/0.021 &78.90/69.16 & 89.02/83.55&97.86/96.84\\ \hline
LIDF &0.082/0.082 &0.183/0.184 &0.069/0.069 &23.70/23.69 &42.77/42.88 &75.44/75.54\\ \hline
Ours &\textbf{0.022/0.025} &\textbf{0.034/0.044} &\textbf{0.013/0.017} &\textbf{81.90/75.27} &\textbf{92.18/89.15} &\textbf{98.39/97.81}\\ \hline

& \multicolumn{6}{c}{STD-CatKnown (Real)} \\ \hline
NLSPN & 0.114/0.047 &0.027/0.031 & 0.015/0.018&94.83/89.47 &98.37/97.48 &99.38/99.32\\ \hline
LIDF &0.019/0.022 &0.019/0.023 &0.013/0.015 &93.08/90.32 &98.39/97.38 &99.83/99.62\\ \hline
Ours &\textbf{0.015/0.018} &\textbf{0.013/0.016} &\textbf{0.008/0.011} &\textbf{96.66/94.97} &\textbf{99.03/98.79} &\textbf{99.92/99.85}\\ \hline

& \multicolumn{6}{c}{STD-CatNovel (Real)} \\ \hline
NLSPN &0.087 & 0.050&0.025 &\textbf{81.95}  &90.36  &96.06 \\ \hline
LIDF & 0.041 & 0.060 & 0.031 & 53.69 & 79.80 & 99.63 \\ \hline
Ours & \textbf{0.025} & \textbf{0.033} & \textbf{0.017} & 81.55 & \textbf{93.10} & \textbf{99.84} \\ \hline
\end{tabular}
\end{center}
%\vspace{-10mm}
\end{table}

\begin{table}
\scriptsize
\renewcommand{\arraystretch}{1.2}
\begin{center}
%%\vspace{-3mm}
\caption{\textbf{Quantitative results for Sim-to-Real.} \emph{Synthetic} means taking the cropped synthetic depth images for training, and \emph{Simulated} means taking the simulated depth images from the train split of DREDS-CatKnown for training.}
%%\vspace{-3mm}
\label{table:depth_restoration_sim2real}
\begin{tabular}{c|cccccc}
\hline
Trainset & RMSE$\downarrow$ & REL$\downarrow$ & MAE$\downarrow$ & $\delta_{1.05}\uparrow$ & $\delta_{1.10}\uparrow$ & $\delta_{1.25}\uparrow$ \\ \hline
& \multicolumn{6}{c}{STD-CatKnown (Real)} \\ \hline
Synthetic & 0.0467/0.056 & 0.0586/0.070 & 0.0377/0.047 & 49.12/39.42 & 86.50/79.85 & 98.98/97.66 \\ \hline
Simulated & \textbf{0.015/0.018} &\textbf{0.013/0.016} &\textbf{0.008/0.011} &\textbf{96.66/94.97} &\textbf{99.03/98.79} &\textbf{99.92/99.85}\\ \hline
& \multicolumn{6}{c}{STD-CatNovel (Real)} \\ \hline
Synthetic & 0.065 & 0.101 & 0.053 & 21.04 & 55.87 & 96.96\\ \hline
Simulated & \textbf{0.025} & \textbf{0.033} & \textbf{0.017} & \textbf{81.55} & \textbf{93.10} & \textbf{99.84} \\ \hline
\end{tabular}
\end{center}
%\vspace{-6mm}
\end{table}

%\vspace{-6mm}
\begin{table}
\scriptsize
\renewcommand{\arraystretch}{1.2}
\begin{center}
\caption{\textbf{Quantitative results for domain transfer.} \emph{The previous best results} means that the best previous method is trained on ClearGrasp and Omniverse, and evaluated on ClearGrasp. \emph{Domain transfer} means that SwinDRNet is trained on DREDS-CatKnown and evaluated on ClearGrasp.}
\label{table:depth_restoration_domain_transfer}
\begin{tabular}{c|cccccc}
\hline
Model & RMSE$\downarrow$ & REL$\downarrow$ & MAE$\downarrow$ & $\delta_{1.05}\uparrow$ & $\delta_{1.10}\uparrow$ & $\delta_{1.25}\uparrow$ \\ \hline
& \multicolumn{6}{c}{ClearGrasp real-known} \\ \hline
The previous best results & 0.028 & 0.033 & 0.020 & 82.37 & 92.98 & 98.63 \\ \hline
Domain transfer & \textbf{0.022} & \textbf{0.017} & \textbf{0.012} & \textbf{91.46} & \textbf{97.47} & \textbf{99.86} \\ \hline

& \multicolumn{6}{c}{ClearGrasp real-novel} \\ \hline
The previous best results & 0.025 & 0.036 & 0.020 & 79.5 & 94.01 & 99.35 \\ \hline
Domain transfer & \textbf{0.016} & \textbf{0.008} & \textbf{0.005} & \textbf{96.73} & \textbf{98.83} & \textbf{99.78} \\ \hline
\end{tabular}
\end{center}
%\vspace{-6mm}
\end{table}
%%\vspace{-8mm}

\textbf{Sim-to-Real and Domain Transfer.}
We perform sim-to-real and domain transfer experiments to verify the generalization ability of the DREDS dataset. For sim-to-real experiments, SwinDRNet is trained on DREDS-CatKnown, but takes different depth images as input of training (one follow~\cite{zhu2021rgb} and takes the cropped synthetic depth image as input, and another takes the simulated depth image). The results evaluated on STD in Table \ref{table:depth_restoration_sim2real} reveal the powerful potential of our depth simulation pipeline, which can significantly close the sim-to-real gap and generalize to the new categories. For domain transfer experiments, we train SwinDRNet on the train split of DREDS-CatKnown dataset and evaluate on Cleargrasp dataset. The results reported in Table \ref{table:depth_restoration_domain_transfer} testify that model only trained on DREDS-CatKnown can easily generalize to the new domain ClaerGrasp and outperform the previous results directly trained on ClearGrasp and Omniverse~\cite{zhu2021rgb} (LIDF train on Omniverse and ClearGrasp), which verifies the generalization ability of our dataset.
% %\vspace{-6mm}

%%\vspace{2mm}
\subsection{Category-level Pose Estimation}
% %\vspace{-2mm}
\textbf{Evaluation Metrics.}
We use two aspects of metrics to evaluate: 1) \textbf{3D IoU.} It computes the intersection over union of ground truth and predicted 3D bounding boxes. We choose the threshold of 25$\%$ (IoU25), 50$\%$(IoU50) and 75$\%$(IoU75) for this metric. 2) \textbf{Rotation and translation errors.} It computes the rotation and translation errors between the ground truth pose and predicted pose. We choose 5$^{\circ}$2cm, 5$^{\circ}$5cm, 10$^{\circ}$2cm, 10$^{\circ}$5cm, 10$^{\circ}$10cm for this metric.

\textbf{Baselines.} We choose two models as baselines to show the usefulness of the restored depth for category-level pose estimation and the effectiveness of SwinDRNet+NOCSHead: 1) \textbf{NOCS}~\cite{wang2019normalized}. It takes a RGB image as input to predict the per-pixel normalized coordinate map and obtain the pose with the help of the depth map. 2) \textbf{SGPA}~\cite{chen2021sgpa}. The state-of-the-art method. It leverages one object and its corresponding category prior to dynamically adapting the prior to the observed object. Then the prior adaptation is used to reconstruct the 3D canonical model of the specific object for pose fitting. 

\textbf{Results.} To verify the usefulness of the restored depth, we report the results of three methods using raw or restored (output of SwinDRNet) depth in Table \ref{table:pose_estimation}. \emph{-only} means using raw depth in the whole experiment, \emph{Refined depth+} means using restored depth for pose fitting in NOCS and SwinDRNet+NOCSHead. Due to the fact that SGPA deforms the point cloud to get the results which are sensitive to depth, we use restored depth for both training and inference. We observe that restored depth improves the performance of three methods by large margins under all the metrics on both dataset. These performance gains suggest that depth restoration is truly useful for category-level pose estimation. Moreover, SwinDRNet+NOCSHead outperforms NOCS and SGPA under all the metrics.

\begin{table}
\scriptsize
\renewcommand{\arraystretch}{1.2}
\begin{center}
\caption{\textbf{Quantitative results for category-level pose estimation.} \emph{only} means using raw depth in the whole experiment,\emph{Refined} means using restored depth for training and inference in SGPA and for pose fitting in NOCS and our method.}%(ACC)}
\label{table:pose_estimation}
\begin{tabular}{c|c c c c c c c c}
\hline
Methods & IoU25 & IoU50 & IoU75& $5^{\circ}2$cm & $5^{\circ}5$cm & $10^{\circ}2$cm & $10^{\circ}5$cm & $10^{\circ}10$cm \\
\hline
& \multicolumn{8}{c}{DREDS-CatKnown (Sim)} \\ \hline
% NOCS-only & 85.7 & 66.0 & 23.0 &21.3 & 25.4 & 40.0 & 47.9 & 49.0\\ \hline
% SGPA-only & 79.5 & 66.7 & 49.1&29.5 & 32.5 & 48.7 & 54.7 & 55.7\\ \hline
% Refined depth + NOCS & 86.7 & 73.2 & 40.7 & 30.4 & 31.8 & 54.1 & 57.5 & 57.6  \\ \hline
% Refined depth + SGPA & 82.3 & 72.0 & 60.5 & 45.9 & 46.8 & 66.4 & 68.4& 68.5 \\ \hline
% Ours-only & 94.3 &82.5 & 57.9& 34.5  & 37.6 & 55.7 & 62.6 & 63.2\\ \hline 
% Refined depth + Ours & \textbf{94.7} &\textbf{84.8} &\textbf{68.0} & \textbf{49.1}  & \textbf{50.1} & \textbf{69.8} & \textbf{72.4} & \textbf{72.5}\\ \hline
NOCS-only & 85.4 & 61.1 & 18.3 & 22.8 & 27.2 & 43.4 & 51.8 & 52.9\\ \hline
SGPA-only & 77.3 & 63.7 & 30.0 & 30.1 & 33.1 & 49.9 & 55.9 & 56.7\\ \hline
Refined depth + NOCS & 85.4 & 65.9 & 27.6 & 32.1 & 33.5 & 57.3 & 60.9 & 60.9  \\ \hline
Refined depth + SGPA & 82.1 & 73.4 & 45.4 & 46.5 & 47.4 & 67.5 & 69.4 & 69.5 \\ \hline
Ours-only & 94.3 & 78.8 & 36.7 & 34.6  & 37.8 & 55.9 & 62.9 & 63.5\\ \hline 
Refined depth + Ours & \textbf{95.3} &\textbf{85.0} &\textbf{49.9} & \textbf{49.3}  & \textbf{50.3} & \textbf{70.1} & \textbf{72.8} & \textbf{72.8}\\ \hline
& \multicolumn{8}{c}{STD-CatKnown (Real)} \\ \hline
% NOCS-only & 83.2 & 66.9 & 16.9 & 20.4 & 26.0 & 37.9 & 52.5 & 53.5\\ \hline
% SGPA-only & 77.6 & 67.1 & 46.6& 30.0 & 32.3 & 47.7 & 53.3 & 53.9\\ \hline
% Refined depth + NOCS & 82.6 & 72.6 & 35.6 & 28.5 & 30.0 & 54.4 & 57.6 & 57.7  \\ \hline
% Refined depth + SGPA & 78.8 & 71.6 & 62.8 & 49.3 & 49.7 & 70.5 &  71.5 & 71.6 \\ \hline
% Ours-only & \textbf{92.4} & 87.4 & 61.7 &37.9 & 42.6 &57.8 & 70.6 & 71.0\\ \hline
% Refined depth + Ours & \textbf{92.4} &\textbf{88.0}& \textbf{75.9} &\textbf{52.9} & \textbf{53.8} &\textbf{77.1} & \textbf{79.1} & \textbf{79.1}\\ \hline
NOCS-only & 89.1 & 63.7 & 17.2 & 23.0 & 28.9 & 42.1 & 57.4 & 58.2\\ \hline
SGPA-only & 75.2 & 63.1 & 30.5 & 31.9 & 34.3 & 50.3 & 56.0 & 56.5\\ \hline
Refined depth + NOCS & 88.8 & 71.1 & 28.7 & 29.8 & 31.2 & 57.4 & 60.6 & 60.7  \\ \hline
Refined depth + SGPA & 77.2 & 71.6 & 49.0 & 51.1 & 51.5 & 72.8 & 73.7 & 73.7 \\ \hline
Ours-only & \textbf{91.5} & 81.3 & 39.3 & 38.2 & 42.9 & 58.3 & 71.2 & 71.5\\ \hline
Refined depth + Ours & \textbf{91.5} &\textbf{85.7}& \textbf{55.7} &\textbf{53.3} & \textbf{54.1} &\textbf{77.6} & \textbf{79.7} & \textbf{79.7}\\ \hline
\end{tabular}
\end{center}
\end{table}

\subsection{Robotic Grasping}
\textbf{Experiments Setting.} 
We conduct real robot experiments to evaluate the depth restoration performance on robotic grasping tasks. In our physical setup, we use a  7-DoF Panda robot arm from Franka Emika with a parallel-jaw gripper. RealSense D415 depth sensor is mounted on the tripod in front of the arm. We set 6 rounds of table clearing experiments. For each round, 4 to 5 specular and transparent objects are randomly picked from STD objects to construct a cluttered scene. For each trial, the robot arm executes the grasping pose with the highest score, and removes the grasped object until the workspace is cleared, or 10 attempts are reached.

% The goal of the grasping experiments is to grasp the objects in the cluttered scene and place them into a box until the workspace is emptied. 
% In our experiments, 20 specular and transparent objects with multiple categories are randomly selected from the object set of STD dataset. We set 6 rounds of table clearing experiments. 
% For each rounds, we randomly pick 4 to 5 objects to construct a cluttered scene. %We do two experiments on the same scene in order to compare the two baselines. 
% %Before each trial, depth sensor captures the RGBD images of the scene, and then the algorithm performs inference to obtain the grasp candidates. 
% For each trial, the robot arm executes the grasping pose with the highest score, and remove the grasped object until the workspace is cleared, or 10 attempts is reached. 
% %the maximum number of We set the maximum is 10.

\textbf{Evaluation Metrics.} 
Real grasping performance is measured using the following metrics: 1) \textbf{Success Rate}: the ratio of grasped object number and attempt number, 2) \textbf{Completion Rate}: the ratio of successfully removed object number and the original object number in a scene.

\textbf{Baselines.}
We follow the 6-DoF grasping pose prediction network GraspNet-baseline, using the released pretrained model. \emph{GraspNet} means GraspNet-baseline directly takes the captured raw depth as input, while \emph{SwinDRNet+GraspNet} means the network receives the refined point cloud from SwinDRNet that is trained only on DREDS-CatKnown dataset.

\begin{table}
\scriptsize
\renewcommand{\arraystretch}{1.2}
\begin{center}
\caption{\textbf{Results of real robot experiments.} \emph{\#Objects} denotes the sum of grasped object numbers in all rounds.
\emph{\#Attempts} denotes the sum of robotic grasping attempt numbers in all rounds.}
\label{table:grasping_realrobot}
\setlength{\tabcolsep}{3pt}
\begin{tabular}{c|cccc}
\hline
Methods & \#Objects & \#Attempts & Success Rate & Completion Rate\\ \hline
GraspNet & 19 & 49 & 38.78\% & 40\% \\ \hline
SwinDRNet+GraspNet & 25 & 26 & \textbf{96.15\%} & \textbf{100\%}\\ \hline
\end{tabular}
\end{center}
\end{table}

\textbf{Results.} 
Table~\ref{table:grasping_realrobot} reports the performance of real robot experiments. \emph{SwinDRNet+GraspNet} obtains high success rate and completion rate, while \emph{GraspNet} is lower. Without depth restoration, it is difficult for a robot arm to grasp specular and transparent objects due to the severely incomplete and inaccurate raw depth. The proposed SwinDRNet significantly improves the performance of specular and transparent object grasping.

%% file: tex_final/07_conclusion.tex
% In this paper, to mitigate the noisy and missing depths problem of commercial depth sensors, our proposed framework, DREDS, synthesises a large-scale RGBD dataset with realistic sensor noises, so as to close the sim-to-real gap for specular and transparent objects.

% We evaluate our method on depth restoration, category-level pose estimation and robot grasping tasks, the quantitative results, as well as the ablation studies, show the effectiveness of our method.

In this work, we propose a powerful RGBD fusion network, SwinDRNet, for depth restoration. Our proposed framework, DREDS, synthesizes a large-scale RGBD dataset with realistic sensor noises, so as to close the sim-to-real gap for specular and transparent objects. Furthermore, we collect a real dataset STD, for real-world performance evaluation. Evaluations on depth restoration, category-level pose estimation, and object grasping tasks demonstrate the effectiveness of our method.

%% file: supptex_final/08_supp_abstract.tex
\iffalse
In the supplementary material, we present a video for this paper and the following additional sections:
\begin{itemize}
  \item Section~\ref{sec:sec1}. Domain Randomization Details.
  \item Section~\ref{sec:sec2}. Additional Dataset Details.
  \item Section~\ref{sec:sec3}. Additional Experiments and Results.
\end{itemize}
\fi

%\begin{abstract}
In the supplementary material, we present the additional sections for this paper, including domain randomization details, network implementation details, additional experiments and results, and additional dataset details.
%\end{abstract}

%% file: supptex_final/09_supp_dr.tex
In this work, we propose the Domain Randomization-Enhanced Depth Simulation (DREDS) approach, leveraging domain randomization and depth sensor simulation to generate photorealistic RGB images
and simulated depths with realistic sensor noises. Specifically, during the simulated data generation, we perform domain randomization in the following aspects:

\textbf{Scene and Object Setting.} 
% object: shape, number, arrangement
We focus on hand-scale objects and a table-top setting. 
We set the scene into the following two types:
1) Category-aware scenes that mainly utilize  ShapeNetCore~\cite{chang2015shapenet} objects from 7 object categories -- camera, car, airplane, bowl, bottle, can, and mug. We also have some distractor objects from categories of phone, guitar, cap, \emph{etc}. In total, we leverage 1536 objects for training and 265 objects for evaluation. In our simulated scenes, we load a random number of objects ranging from 6 to 10 with random scales and categories and let them fall freely under gravity onto a ground plane to create random but physically plausible spatial arrangements of objects and prepare cluttered scenes. 
2) Category-agnostic scenes. To evaluate the generalization ability to category-novel objects and the performance of grasping, we adopt 60 objects from GraspNet-1Billion~\cite{fang2020graspnet}. We follow their original poses and arrangements but transfer random types of material as described in the next section.

\textbf{Material Modeling and Assignment.}
% material modeling: BSDF
Few of the existing depth sensor simulators consider modeling a variety of randomized real-world materials, especially specular and transparent materials. In this work, we adopt a bidirectional scattering distribution function (BSDF)~\cite{bartell1981theory},  a unified representation covering the most common materials. BSDF defines how the light is scattered on a surface to determine the material of each point on the object.

% diffuse and specular material modeling, transparent material modeling
Specifically, we use Disney principled BSDF~\cite{burley2015extending,burley2012physically} $f_{D\&S}(\phi)$ for diffuse and specular material modeling, where $\phi$ is the set of scalar parameters or nested functions, including the base color, subsurface, metallic, specular, roughness, anisotropic, \emph{etc}. We use a mix of BSDF $f_{T}(\psi)$ to represent transparent materials, containing glass BSDF, transparent BSDF, and translucent BSDF to adjust transparency, as well as refraction BSDF to add refraction, and glossy BSDF to add reflection on the surface, \emph{etc}, where $\psi$ means the parameter set from each BSDF function like surface color, index of refraction (IOR), and roughness.

% asset, parameter randomization
Based on the above BSDF models, we collect an asset of materials with different categories that cover common objects in life, including 1) 27 specular materials including metal, porcelain, clean plastic, paint, \emph{etc.}, 2) 4 transparent materials, 3) 36 diffuse materials including rubber, leather, wood, fabric, coarse plastic, paper, clay, \emph{etc.} We randomize the parameters of the BSDF function for each material within a range, generating a large-scale material collection with wide variations.

% assign to objects, material transfer
We assign one type of material to each object in the scene randomly. For those objects with default colors or texture maps, we mix their colors or textures with the base color of the assigned material in a randomized ratio. It means that we can easily transfer an existing synthetic object dataset to a dataset with a large amount of specular and transparent objects.

\textbf{Camera Setting.}
% camera: location, angle
We follow RealSense D415 to set up the projector’s parameters (\emph{e.g.}, the IR pattern image, baseline distance) and other cameras' intrinsic parameters. Camera locations and poses are randomized within a range, so that the objects in each scene can be captured from arbitrary directions.

\textbf{Lighting and Background Setting.}
% lighting: different environment map, transformation (rotation and translation) of environment map, strength
We collect 74 HDRI environment maps for training, and 23 for testing, including indoor and outdoor scenes, as well as natural and artificial lighting. An arbitrarily chosen environment map with random intensities is used to simulate realistic ambient illumination. 
% background
For the background, we pick 81 common indoor materials for training and 23 for evaluation, including wood, marble, tiles, concrete, \emph{etc}. 
A random selection of these materials is applied to the ground plane to increase variations of the scene.

%% file: supptex_final/11_supp_implement.tex
We implement the proposed SwinDRNet and downstream algorithms in PyTorch. We train SwinDRNet for 20 epochs (nearly 146,000 iterations) with batch size 32, using AdamW~\cite{loshchilov2018decoupled} optimizer with $\beta_1$ = 0.9, $\beta_2$ = 0.999, a learning rate of 1e-4, a weight decay of 0.01, as well as a learning scheduler with a linear warmup of 500 iterations and a linear learning rate decay. SwinDRNet takes RGB and raw depth images that are resized to 224*224 as the input, and outputs the restored depth image with the same size for downstream tasks. Note that for SGPA~\cite{chen2021sgpa}, the baseline method of category-level pose estimation, as its performance depends on the number of points of the input point cloud, i.e., the resolution of the depth, the original RGBD images are firstly resized to 224*448, and then sampled at an interval of 1 along the direction of the row to obtain two 224*224 inputs, as well as two 224*224 outputs from the network. We finally interpolate these two outputs in the same sampling way above, to obtain the 224*448 depth as the input to SGPA.

%% file: supptex_final/12_supp_experiment.tex
\subsection{Depth Restoration}
\begin{figure}[htbp]
% \centering
% \includegraphics[scale=0.38]{figure/depth_restoration.pdf}

\centering % trim = <left low right upper
\includegraphics[trim=0 0 0 0,clip, width=\linewidth]{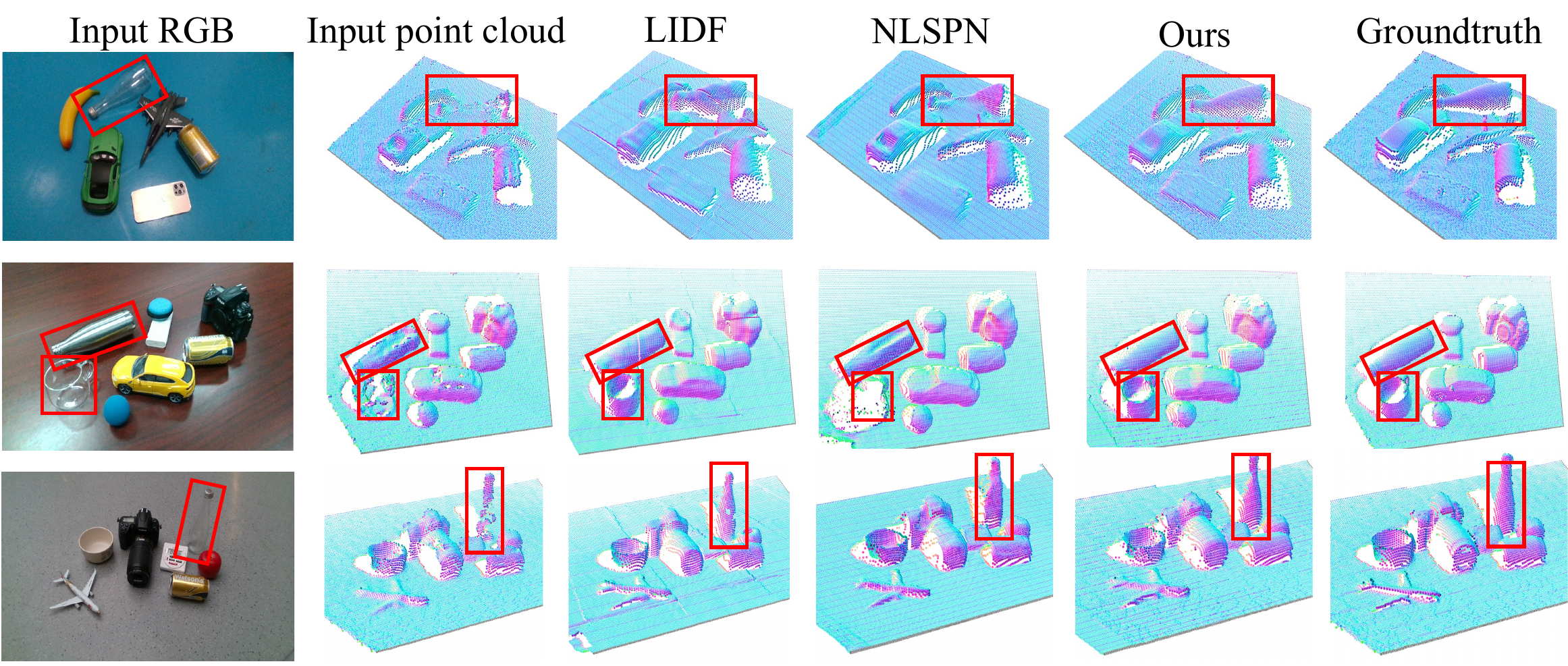}

\caption{\textbf{Qualitative comparison to state-of-the-art methods.} For a fair comparison, all the methods are trained on the train split of DREDS-CatKnown. Red boxes highlight the specular or transparent objects.}
\vspace{-5mm}
\label{fig: Qualitative comparison to state-of-the-art methods}
\end{figure}

\textbf{Qualitative Comparison to State-of-the-art Methods.} 
Figure \ref{fig: Qualitative comparison to state-of-the-art methods} shows the qualitative comparison of STD dataset, demonstrating that our method can predict a more accurate depth on the area with missing or incorrect values while preserving the depth value of the correct area of the raw depth map.

%\vspace{-5mm}

\textbf{Cross-Sensor Evaluation.} In this work, depth sensor simulation and real-world data capture are both based on Intel RealSense D415. To investigate the robustness of the proposed SwinDRNet on other types of depth sensors, we evaluate the performance on data of two scenes from STD-CatKnown dataset, captured by Intel RealSense D435. Table \ref{table:depth_restoration_d435} shows a comparison of the results evaluated on D415 and D435 data after training on DREDS-CatKnown dataset. We observe that SwinDRNet has similar performance on data from these two different depth sensors in each scene, which verifies the good cross-sensor generalization ability of SwinDRNet.

\begin{table}
\scriptsize
\setlength{\tabcolsep}{3pt} % Default value: 6pt
\renewcommand{\arraystretch}{1.2} % Default value: 1
\begin{center}
\vspace{-4mm}
\caption{\textbf{Quantitative results for cross-sensor evaluation.} The performance of SwinDRNet is evaluated on RGB-D data captured by Intel RealSense D415 and D435 in each of the two scenes.}
\label{table:depth_restoration_d435}
%\resizebox{1.\columnwidth}{!}{
\begin{tabular}{cc|cccccc}
\hline
Scenes & Sensors & RMSE$\downarrow$ & REL$\downarrow$ & MAE$\downarrow$ & $\delta_{1.05}\uparrow$ & $\delta_{1.10}\uparrow$ & $\delta_{1.25}\uparrow$ \\ \hline
\multirow{2}{*}{1} & D415 & 0.017/0.017 & 0.015/0.016 & 0.009/0.010 & 94.62/94.30 & 98.34/98.60 & 99.94/99.95 \\ 
 & D435 & 0.021/0.023 & 0.022/0.025 & 0.013/0.015 & 89.30/86.23 & 97.95/97.85 & 99.95/99.98 \\ \hline

\multirow{2}{*}{2} & D415 & 0.013/0.018 & 0.011/0.014 & 0.008/0.011 & 97.93/96.02 & 99.47/98.94 & 100.00/100.00 \\ 
 & D435 & 0.016/0.024 & 0.015/0.024 & 0.010/0.017 & 95.25/89.29 & 99.16/97.69 & 100.00/100.00 \\ \hline
\end{tabular}
%}
\vspace{-5mm}
\end{center}
\end{table}

\vspace{-10mm}
\subsection{Category-level Pose Estimation}
\begin{figure}[htbp]
%\vspace{-5mm}
\centering
\centering % trim = <left low right upper
	\includegraphics[trim=0 0 0 0,clip, width=\linewidth]
{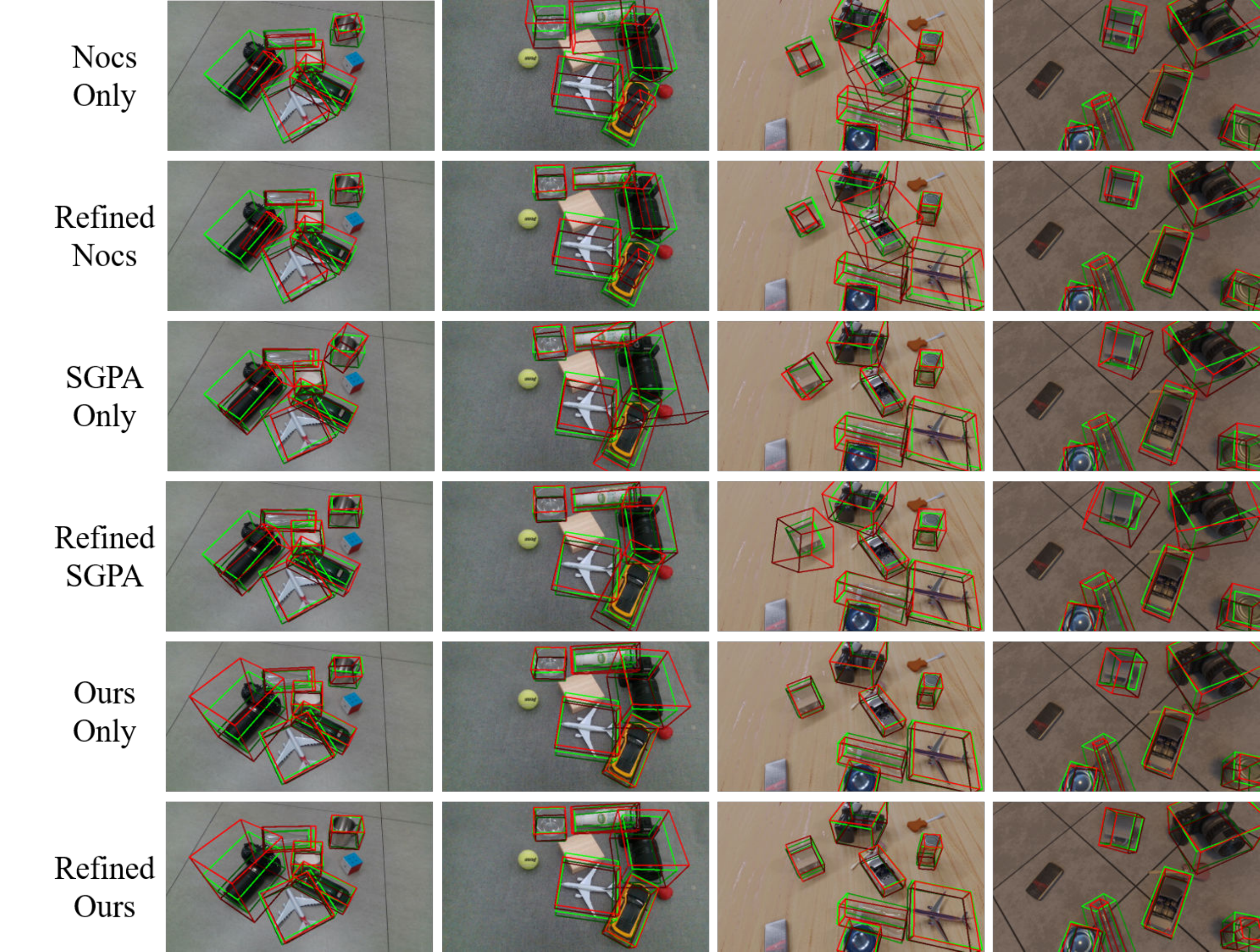}
%\vspace{-3mm}
\caption{\textbf{Qualitative results of pose estimations on DREDS and STD datasets.} The ground truths are shown in green while the estimations are shown in red. \emph{only} means using raw depth in the whole experiment, \emph{Refined} means using restored depth for training and inference in SGPA and for pose fitting in NOCS and our method.}
\label{fig: pose}
\vspace{-10mm}
\end{figure}

\textbf{Qualitative Comparison to Baseline Methods.} 
Figure \ref{fig: pose} shows the qualitative results of different experiments on DREDS and STD datasets. We can see that the qualities of our predictions are generally better than others. The figure also shows that NOCS~\cite{wang2019normalized}, SGPA~\cite{chen2021sgpa} and our method all perform better with the help of restoration depth, especially for specular and transparent objects like the mug, bottle and bowl, which indicates that depth restoration does help category-level pose estimation task.

\textbf{Quantitative Comparison to Restored Depth Inputs.} We further evaluate the influence of different restored depths for category-level pose estimation, which is presented in Table \ref{table:ablation_studies_diff_depth_pose}. The proposed SwinDRNet+NOCSHead network receives the restored depth from SwinDRNet and the competing depth restoration methods for pose fitting. Quantitative results under all metrics demonstrate the superiority of SwinDRNet over other baseline methods in boosting the performance of category-level pose estimation.

\begin{table}
\vspace{-5mm}
\scriptsize
\renewcommand{\arraystretch}{1.2}
\begin{center}
\caption{\textbf{Quantitative results for category-level pose estimation using different restored depths from SwinDRNet and the competing baseline methods.} The left of '/' shows the results evaluated on all objects, and the right of '/' shows the results evaluated on specular and transparent objects.
\vspace{-3mm}}
\label{table:ablation_studies_diff_depth_pose}
\begin{tabular}{c|c c c c c c c c}
\hline%\noalign{\smallskip}
Methods & IoU25 & IoU50 & IoU75 & $5^{\circ}2$cm & $5^{\circ}5$cm & $10^{\circ}2$cm & $10^{\circ}5$cm & $10^{\circ}10$cm \\
%\noalign{\smallskip}
\hline
%\noalign{\smallskip}
& \multicolumn{8}{c}{DREDS-CatKnown (Sim)} \\ \hline
% NLSPN & \textbf{94.7}/98.1 & 84.6/90.3 & 65.9/71.2 &39.4/39.4 & 40.3/40.4 & 65.2/67.8& 67.6/70.4 & 67.6/70.4  \\ \hline
% LIDF & 94.4/97.9 & 83.3/89.5 & 59.3/66.4 & 33.7/37.4 & 36.3/39.8& 57.9/63.7& 64.3/69.8 & 64.6/70.0\\ \hline
% Ours& \textbf{94.7}/\textbf{98.2}&\textbf{84.8}/\textbf{90.8}& \textbf{68.0}/\textbf{74.0}&\textbf{49.1}/\textbf{51.5} & \textbf{50.1}/\textbf{52.9} & \textbf{69.8}/\textbf{73.9} & \textbf{72.4}/\textbf{77.0} & \textbf{72.5}/\textbf{77.1}\\ \hline
NLSPN & 95.1/97.5 & 83.8/87.4 & 46.4/48.9 & 39.6/39.6 & 40.4/40.5 & 65.5/68.1 & 67.9/70.7 & 67.9/70.8  \\ \hline
LIDF & 94.6/97.1 & 80.7/85.0 & 36.6/40.8 & 33.9/37.5 & 36.4/40.0 & 58.2/64.0 & 64.7/70.2 & 64.9/70.4\\ \hline
Ours& \textbf{95.3}/\textbf{97.8}&\textbf{85.0}/\textbf{88.9}& \textbf{49.9}/\textbf{52.4}&\textbf{49.3}/\textbf{51.8} & \textbf{50.3}/\textbf{53.1} & \textbf{70.1}/\textbf{74.3} & \textbf{72.8}/\textbf{77.4} & \textbf{72.8}/\textbf{77.5}\\ \hline
& \multicolumn{8}{c}{STD-CatKnown (Real)} \\ \hline
% NLSPN & 92.3/99.5 & 87.7/94.8 & 73.5/73.5 & 45.2/31.5 & 46.2/33.3 & 72.5/57.1 &75.1/60.9 & 75.1/60.9 \\ \hline
% LIDF &92.3/99.1& 87.2/93.4 & 67.0/68.5 &34.6/35.4& 37.1/40.2& 64.7/60.8 & 70.4/\textbf{69.0} &  70.5/\textbf{69.2}\\ \hline
% Ours& \textbf{92.4}/\textbf{99.7}&\textbf{88.0}/\textbf{95.0} &\textbf{ 75.9}/\textbf{78.8} &  \textbf{52.9}/\textbf{40.0}  & \textbf{53.8}/\textbf{41.3} & \textbf{77.1}/\textbf{66.3} & \textbf{79.1}/68.7& \textbf{79.1}/68.7\\ \hline 
NLSPN & 91.4/\textbf{97.2} & 85.2/89.4 & 53.6/48.8 & 45.5/31.6 & 46.5/33.4 & 73.1/57.2 & 75.7/61.1 & 75.7/61.1 \\ \hline
LIDF & 91.3/96.7 & 83.2/85.5 & 42.9/35.9 & 34.8/35.5 & 37.4/40.3 & 65.2/61.1 & 71.0/\textbf{69.3} & 71.1/\textbf{69.4}\\ \hline
Ours& \textbf{91.5}/97.1 &\textbf{85.7}/\textbf{89.9} &\textbf{55.7}/\textbf{54.4} &  \textbf{53.3}/\textbf{40.1}  & \textbf{54.1}/\textbf{41.4} & \textbf{77.6}/\textbf{66.5} & \textbf{79.7}/68.8 & \textbf{79.7}/68.9\\ \hline 
\end{tabular}
\end{center}
\vspace{-8mm}
\end{table}

\vspace{-5mm}
\subsection{Robotic Grasping}
\begin{figure}[htbp]
\vspace{-2mm}
\centering
\includegraphics[scale=0.25]{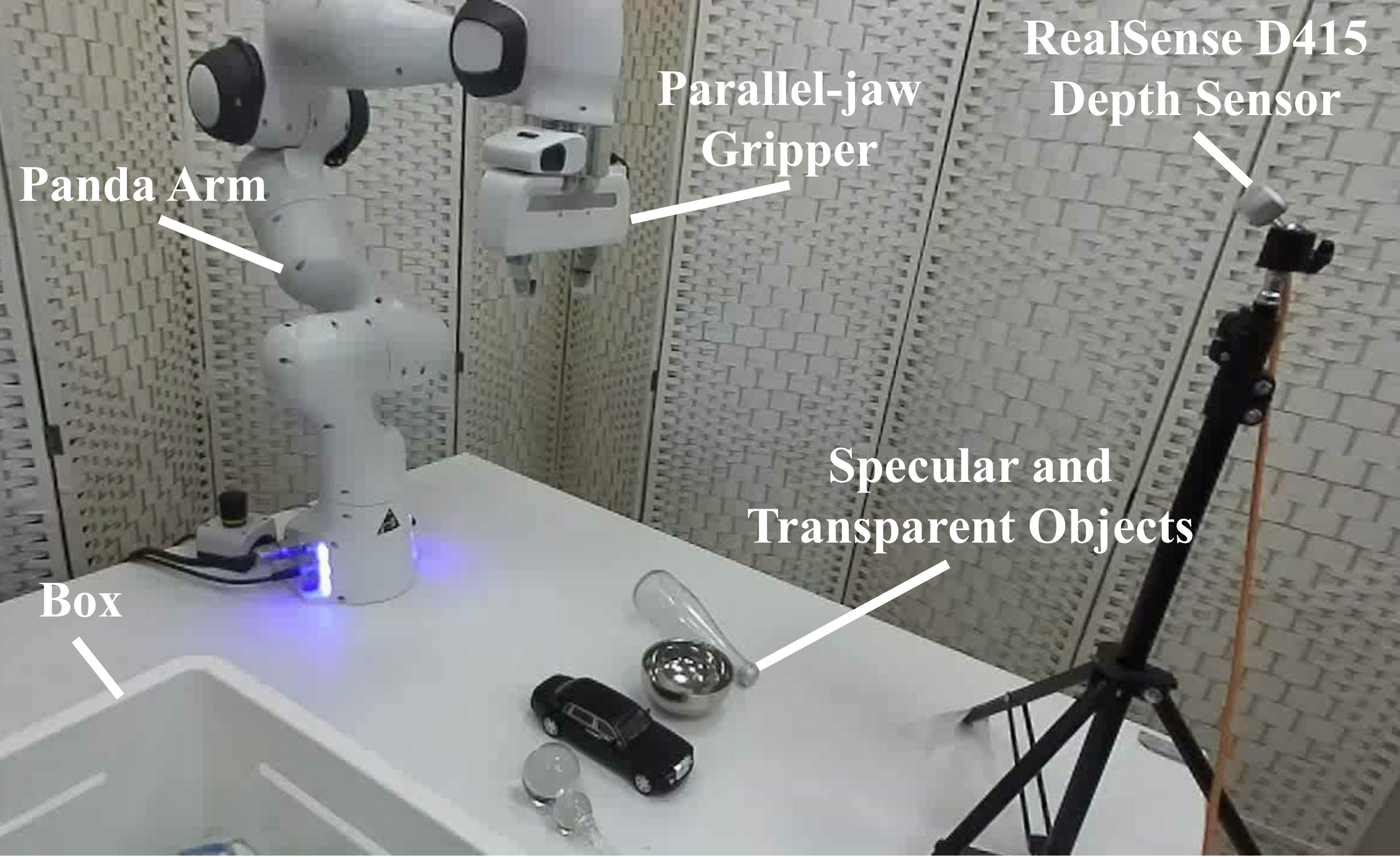}
%\vspace{-5mm}
\caption{\textbf{The setting of real robot experiment for specular and transparent object grasping.}}
%\vspace{-8mm}
\label{fig: robot}
\end{figure}

The illustration of a real robot experiment for specular and transparent object grasping is shown in Figure \ref{fig: robot}. We carry out the table-clearing using the Franka Emika Panda robot arm with the parallel-jaw gripper, and RealSense D415 depth sensor for RGBD images capture.

\vspace{-3mm}
\subsection{Ablation Study}
To analyze the components of the proposed SwinDRNet, as well as domain randomization and the scale of the proposed DREDS dataset, we conduct the ablation studies with different configurations.

\textbf{Analysis of the Modules of SwinDRNet.} We first evaluate the effect of different modules of SwinDRNet with three configurations: 1) Take the concatenated RGBD images as input without the RGB-D fusion and confidence interpolation module. 2) Have no confidence module compared with SwinDRNet. 3) The complete SwinDRNet. As shown in Table \ref{table:ablation_studies_modules}, the performance of depth restoration improves when using these two modules. Note that the network with and without the confidence interpolation module obtain the similar depth restoration performance. However, in Table~\ref{table:ablation_studies_modules_for_pose_estimation}, we observe that SwinDRNet with this module achieves higher performance on object pose estimation, because the module keeps the correct geometric features from the original depth input which benefits the downstream task. The results above indicate the effectiveness of the RGB-D fusion and confidence interpolation module of SwinDRNet.

\textbf{Analysis of Material Randomization.} We analyze the effect of material randomization on depth restoration. We create a dataset of the same size as the fully randomized DREDS-CatKnown dataset. The original materials from ShapeNetCore~\cite{chang2015shapenet} are directly applied to the objects without any transfer or randomization of specular, transparent, diffuse materials.
Table \ref{table:mat_random} shows the results of depth restoration, evaluating on specular and transparent objects. Without material randomization, the performance drops significantly, since the network cannot consider real-world data as the variation of the synthetic training data without seeing sufficient material variation, which demonstrates the significance of material randomization.

\textbf{Analysis of the Scale of Training Data.} In Table \ref{table:ablation_scale}, we show the performance dependence on the dataset scale. Compared to the full scale, the depth restoration performance of SwinDRNet trained on the half scale also degraded, demonstrating the necessity of the scale of the DREDS dataset for the method.

%\vspace{-5mm}
\begin{table}
\scriptsize
\renewcommand{\arraystretch}{1.2}
\begin{center}
\caption{\textbf{Ablation studies for the effect of different modules on depth restoration.} \checkmark denotes prediction with the module.
\vspace{-3mm}
}
\label{table:ablation_studies_modules}
\begin{tabular}{c c|c c c c c c}
\hline%\noalign{\smallskip}
Fusion & Confidence & RMSE$\downarrow$ & REL$\downarrow$ & MAE$\downarrow$ & $\delta_{1.05}\uparrow$ & $\delta_{1.10}\uparrow$ & $\delta_{1.25}\uparrow$ \\
%\noalign{\smallskip}
\hline
%\noalign{\smallskip}
& & \multicolumn{6}{c}{STD-CatKnown} \\ \hline
 & & 0.019/0.027 & 0.019/0.032 & 0.0123/0.021 & 91.09/79.20 & 98.92/97.73 & \textbf{99.95}/\textbf{99.91}\\ \hline
\checkmark & & \textbf{0.014}/\textbf{0.017} & \textbf{0.013}/0.017 & 0.009/0.012 & 96.33/94.18 & \textbf{99.36}/\textbf{99.01} & 99.92/\textbf{99.91}\\ \hline
\checkmark & \checkmark & 0.015/0.018 &\textbf{0.013}/\textbf{0.016} &\textbf{0.008}/\textbf{0.011} &\textbf{96.66}/\textbf{94.97} &99.03/98.79 & 99.92/99.85\\ \hline
\end{tabular}
\end{center}
\end{table}

\vspace{-10mm}
\begin{table}
\scriptsize
\renewcommand{\arraystretch}{1.2}
\begin{center}
\vspace{-6mm}
\caption{\textbf{The effect of confidence for category-level pose estimation.}
\vspace{-3mm}}
\label{table:ablation_studies_modules_for_pose_estimation}
\begin{tabular}{c|c c c c c c c c}
\hline%\noalign{\smallskip}
Confidence & IoU25 & IoU50 & IoU75 & $5^{\circ}$2cm & $5^{\circ}5$cm & $10^{\circ}2$cm & $10^{\circ}5$cm & $10^{\circ}10$cm \\
\hline
& \multicolumn{8}{c}{STD-CatKnown (Real)} \\ \hline
%  & \textbf{92.4} & \textbf{88.0} & 75.6& 51.0 & 51.9 & 76.0 & 78.2 & 78.3\\ \hline
% \checkmark & \textbf{92.4} &\textbf{88.0} &\textbf{75.9}&\textbf{52.9} & \textbf{53.8} &\textbf{77.1} & \textbf{79.1} & \textbf{79.1}\\ \hline
 & \textbf{91.5} & \textbf{85.7} & \textbf{56.2} & 51.3 & 52.2 & 76.6 & 78.8 & 78.8\\ \hline
\checkmark & \textbf{91.5} & \textbf{85.7}& 55.7 &\textbf{53.3} & \textbf{54.1} &\textbf{77.6} & \textbf{79.7} & \textbf{79.7}\\ \hline
\end{tabular}
\end{center}
\vspace{-6mm}
\end{table}

\vspace{-10mm}
\begin{table}
\scriptsize
\renewcommand{\arraystretch}{1.2}
\begin{center}
\caption{\textbf{Quantitative results for material randomization on depth restoration task.} The left of '/' shows the results evaluated on all objects, and the right of '/' evaluated on specular and transparent objects. Note that only one result is reported on STD-CatNovel, because all the objects are specular or transparent.
\vspace{-3mm}}
\label{table:mat_random}
\resizebox{1.\columnwidth}{!}{
\begin{tabular}{c|cccccc}
\hline
Model & RMSE$\downarrow$ & REL$\downarrow$ & MAE$\downarrow$ & $\delta_{1.05}\uparrow$ & $\delta_{1.10}\uparrow$ & $\delta_{1.25}\uparrow$ \\ \hline
& \multicolumn{6}{c}{STD-CatKnow (Real)} \\ \hline
Fixed material & 0.024/0.038 & 0.024/0.045 & 0.015/0.029 & 86.20/65.63 & 96.12/90.94 & 99.87/99.72 \\ \hline
Full randomization &\textbf{0.015/0.018} &\textbf{0.013/0.016} &\textbf{0.008/0.011} &\textbf{96.66/94.97} &\textbf{99.03/98.79} &\textbf{99.92/99.85} \\ \hline
& \multicolumn{6}{c}{STD-CatNovel (Real)} \\ \hline
Fixed material & 0.038 & 0.051 & 0.027 & 67.52 & 84.86 & 98.51 \\ \hline
Full randomization & \textbf{0.025} & \textbf{0.033} & \textbf{0.017} & 81.55 & \textbf{93.10} & \textbf{99.84} \\ \hline
\end{tabular}
}
\end{center}
\vspace{-8mm}
\end{table}

\begin{table}
\scriptsize
\renewcommand{\arraystretch}{1.2}
\begin{center}
%\vspace{-20mm}
\caption{\textbf{Ablation study for the scale of training data on depth restoration.} SwinDRNet is trained on DREDS-CatKnown and evaluated on the specular and transparent objects of STD.
%\vspace{-3mm}
}
\begin{tabular}{c|cccccc}
\hline
Scale & RMSE$\downarrow$ & REL$\downarrow$ & MAE$\downarrow$ & $\delta_{1.05}\uparrow$ & $\delta_{1.10}\uparrow$ & $\delta_{1.25}\uparrow$ \\ \hline
& \multicolumn{6}{c}{STD-CatKnow (Real)} \\ \hline
Half & 0.021 & 0.020 & 0.014 & 92.71 & 98.54 & 99.83 \\\hline
Full &\textbf{0.018}&\textbf{0.016}&\textbf{0.011}&\textbf{94.97}&\textbf{98.79}&\textbf{99.84}\\ \hline
& \multicolumn{6}{c}{STD-CatNovel (Real)} \\ \hline
Half & 0.028 & 0.037 & 0.020 & 80.37 & 91.16 & 99.79 \\ \hline
Full & \textbf{0.025} & \textbf{0.033} & \textbf{0.017} & \textbf{81.55} & \textbf{93.10} & \textbf{99.84} \\ \hline
\end{tabular}
\label{table:ablation_scale}
\end{center}
%\vspace{-8mm}
\end{table}

%% file: supptex_final/10_supp_dataset.tex
\subsection{DREDS Dataset}
\begin{figure}[htbp]
\centering
\centering % trim = <left low right upper
	\includegraphics[trim=0 0 0 0,clip, width=\linewidth]
{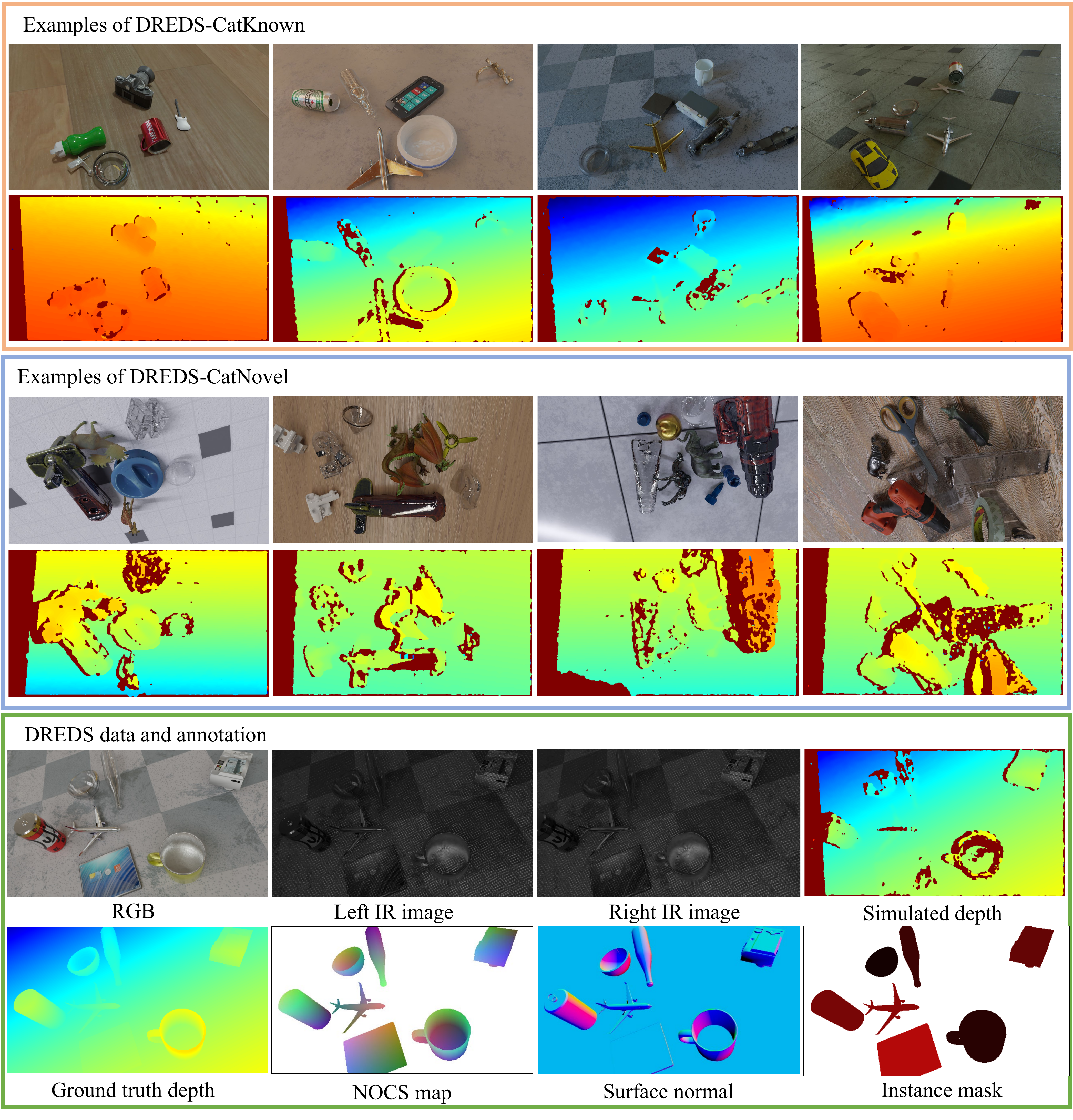}
\caption{\textbf{Paired RGB and simulated depth examples and annotations of DREDS-CatKnown and DREDS-CatNovel datasets.}}
\label{fig: DREDS_example}
\end{figure}

\vspace{-2mm}
We present the DREDS-CatKnown dataset, where the category-level objects are from ShapeNetCore~\cite{chang2015shapenet}, and the DREDS-CatNovel dataset, where we transfer random materials to the objects of GraspNet-1Billion~\cite{fang2020graspnet}. Figure \ref{fig: DREDS_example} shows the examples and annotations of DREDS dataset. For each virtual scene, we provide the RGB image, stereo IR images, simulated depth, ground truth depth, NOCS map, surface normal, instance mask, \emph{etc}.

\subsection{STD Dataset}
\begin{figure}[htbp]
\centering
\centering % trim = <left low right upper
	\includegraphics[trim=0 0 0 0,clip, width=\linewidth]
{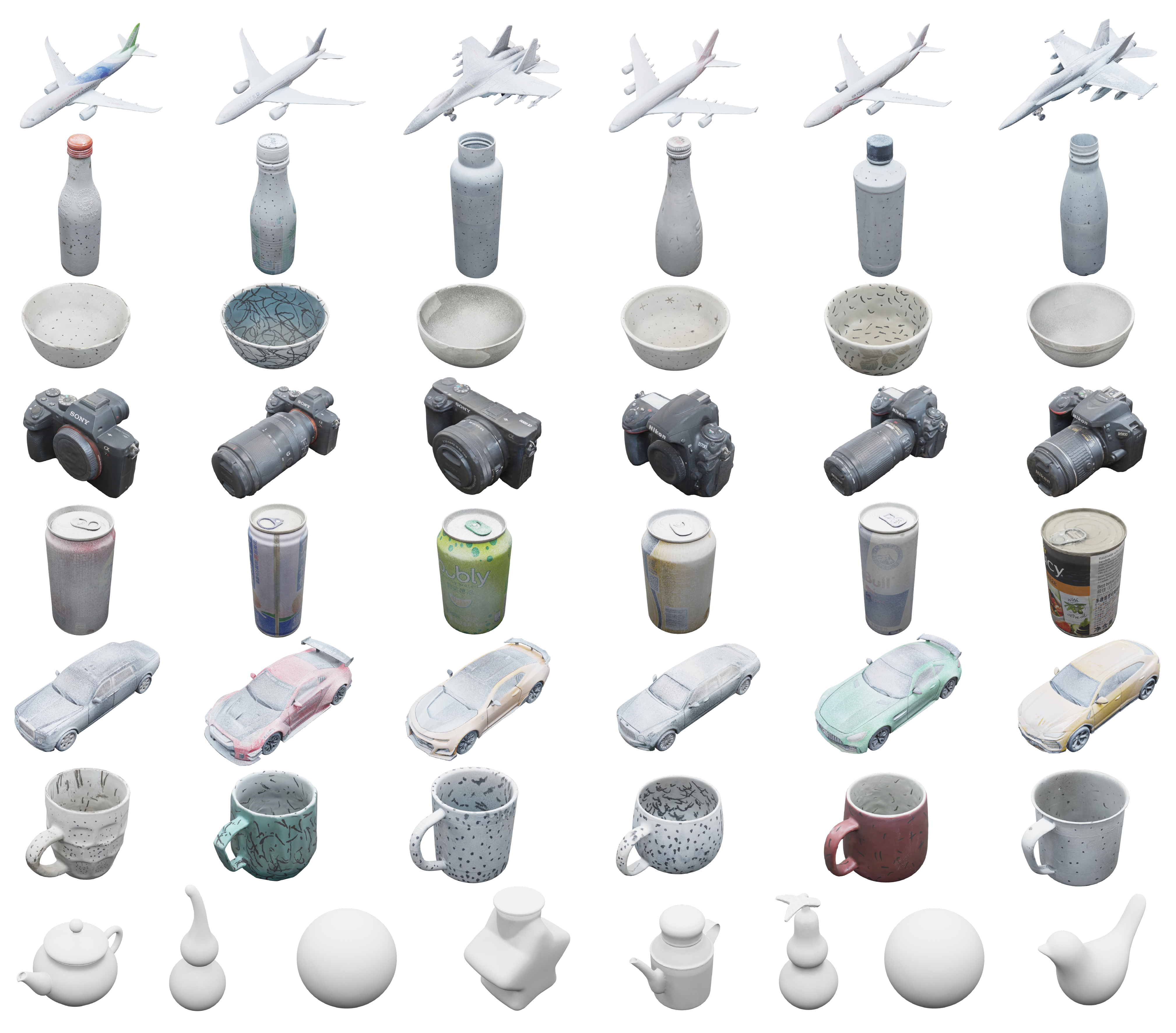}
\vspace{-6mm}
\caption{\textbf{CAD models of the STD object set.} The 1st to 7th
rows show 42 objects in 7 categories, and the last row shows 8 objects in novel categories. }
\label{fig: cad_model}
\vspace{-5mm}
\end{figure}

\vspace{-1mm}
\textbf{Example of CAD Models.} We obtain CAD models of 42 category-level objects and 8 category-novel objects using the 3D reconstruction algorithm. For most of the objects, especially specular and transparent objects, we spray the dye and decorate objects with ink to enhance the reconstruction performance. 50 CAD models are shown in Figure \ref{fig: cad_model}.

\vspace{-6mm}

\begin{figure}[htbp]
\centering
\centering % trim = <left low right upper
	\includegraphics[trim=0 0 0 0,clip, width=\linewidth]
{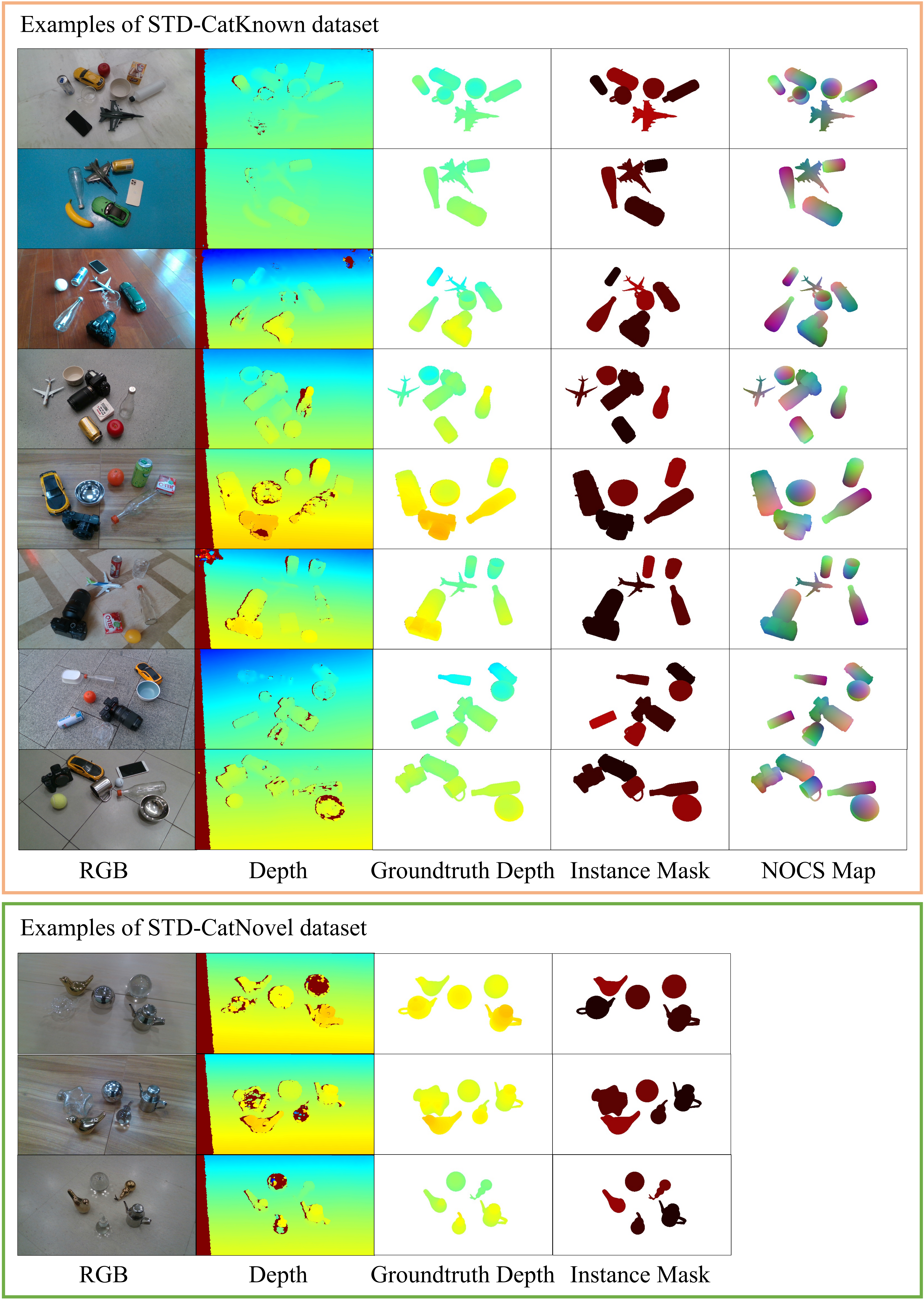}
%\vspace{-5mm}
\caption{\textbf{Examples and annotations of STD-CatKnown and STD-CatNovel datasets.} The ground truth depth maps are labeled only in the area of 42 objects in 7 categories and 8 objects in novel categories. Moreover, the NOCS maps are not annotated in STD-CatNovel dataset because we do not define the normalized object coordinate space for novel categories.}
\label{fig: STD_example}
\vspace{-6mm}
\end{figure}

\textbf{Data Annotation.} 
It is quite time-consuming to annotate such a large amount of real data. We propose to annotate the 6D poses of the objects in the first frame of each scene. Then the annotated 6D poses are propagated to the subsequent frames according to the camera poses with respect to the first frame. We calculate the camera poses using COLMAP~\cite{schoenberger2016sfm}. In our annotation, we develop a program with GUI, enabling the user to move the CAD model, switching back and forth between the 2D image and 3D point cloud space to determine its pose, which facilitates labeling specular and transparent objects whose point clouds are severely missing or incorrect. After annotating 6D pose, we can easily render other annotations like the ground truth depth, instance mask, \emph{etc}. Figure \ref{fig: STD_example} shows the examples and annotations of DREDS dataset. 

%% file: main_final.bbl
\begin{thebibliography}{10}
\providecommand{\url}[1]{\texttt{#1}}
\providecommand{\urlprefix}{URL }
\providecommand{\doi}[1]{https://doi.org/#1}

\bibitem{Blender}
Blender. https://www.blender.org/

\bibitem{macOSAPI}
Object capture api on macos.
  https://developer.apple.com/augmented-reality/object-capture/

\bibitem{bartell1981theory}
Bartell, F.O., Dereniak, E.L., Wolfe, W.L.: The theory and measurement of
  bidirectional reflectance distribution function (brdf) and bidirectional
  transmittance distribution function (btdf). In: Radiation scattering in
  optical systems. vol.~257, pp. 154--160. SPIE (1981)

\bibitem{breyer2020volumetric}
Breyer, M., Chung, J.J., Ott, L., Roland, S., Juan, N.: Volumetric grasping
  network: Real-time 6 dof grasp detection in clutter. In: Conference on Robot
  Learning (2020)

\bibitem{burley2015extending}
Burley, B.: Extending the disney brdf to a bsdf with integrated subsurface
  scattering. Physically Based Shading in Theory and Practice’SIGGRAPH Course
   (2015)

\bibitem{burley2012physically}
Burley, B., Studios, W.D.A.: Physically-based shading at disney. In: ACM
  SIGGRAPH. vol.~2012, pp.~1--7. vol. 2012 (2012)

\bibitem{calli2017yale}
Calli, B., Singh, A., Bruce, J., Walsman, A., Konolige, K., Srinivasa, S.,
  Abbeel, P., Dollar, A.M.: Yale-cmu-berkeley dataset for robotic manipulation
  research. The International Journal of Robotics Research  \textbf{36}(3),
  261--268 (2017)

\bibitem{chang2015shapenet}
Chang, A.X., Funkhouser, T., Guibas, L., Hanrahan, P., Huang, Q., Li, Z.,
  Savarese, S., Savva, M., Song, S., Su, H., et~al.: Shapenet: An
  information-rich 3d model repository. arXiv preprint arXiv:1512.03012  (2015)

\bibitem{chen2021sgpa}
Chen, K., Dou, Q.: Sgpa: Structure-guided prior adaptation for category-level
  6d object pose estimation. In: Proceedings of the IEEE/CVF International
  Conference on Computer Vision. pp. 2773--2782 (2021)

\bibitem{eigen2014depth}
Eigen, D., Puhrsch, C., Fergus, R.: Depth map prediction from a single image
  using a multi-scale deep network. Advances in neural information processing
  systems  \textbf{27} (2014)

\bibitem{fang2020graspnet}
Fang, H.S., Wang, C., Gou, M., Lu, C.: Graspnet-1billion: A large-scale
  benchmark for general object grasping. In: Proceedings of the IEEE/CVF
  conference on computer vision and pattern recognition. pp. 11444--11453
  (2020)

\bibitem{he2020pvn3d}
He, Y., Sun, W., Huang, H., Liu, J., Fan, H., Sun, J.: Pvn3d: A deep point-wise
  3d keypoints voting network for 6dof pose estimation. In: Proceedings of the
  IEEE/CVF conference on computer vision and pattern recognition. pp.
  11632--11641 (2020)

\bibitem{hu2021penet}
Hu, M., Wang, S., Li, B., Ning, S., Fan, L., Gong, X.: Penet: Towards precise
  and efficient image guided depth completion. In: 2021 IEEE International
  Conference on Robotics and Automation (ICRA). pp. 13656--13662. IEEE (2021)

\bibitem{jiang2021synergies}
Jiang, Z., Zhu, Y., Svetlik, M., Fang, K., Zhu, Y.: Synergies between
  affordance and geometry: 6-dof grasp detection via implicit representations.
  In: Robotics: Science and Systems XVII, Virtual Event, July 12-16, 2021
  (2021)

\bibitem{jiao2018look}
Jiao, J., Cao, Y., Song, Y., Lau, R.: Look deeper into depth: Monocular depth
  estimation with semantic booster and attention-driven loss. In: Proceedings
  of the European conference on computer vision (ECCV). pp. 53--69 (2018)

\bibitem{khirodkar2019domain}
Khirodkar, R., Yoo, D., Kitani, K.: Domain randomization for scene-specific car
  detection and pose estimation. In: 2019 IEEE Winter Conference on
  Applications of Computer Vision (WACV). pp. 1932--1940. IEEE (2019)

\bibitem{landau2015simulating}
Landau, M.J., Choo, B.Y., Beling, P.A.: Simulating kinect infrared and depth
  images. IEEE transactions on cybernetics  \textbf{46}(12),  3018--3031 (2015)

\bibitem{liu2021swin}
Liu, Z., Lin, Y., Cao, Y., Hu, H., Wei, Y., Zhang, Z., Lin, S., Guo, B.: Swin
  transformer: Hierarchical vision transformer using shifted windows. In:
  Proceedings of the IEEE/CVF International Conference on Computer Vision. pp.
  10012--10022 (2021)

\bibitem{long2021adaptive}
Long, X., Lin, C., Liu, L., Li, W., Theobalt, C., Yang, R., Wang, W.: Adaptive
  surface normal constraint for depth estimation. In: Proceedings of the
  IEEE/CVF International Conference on Computer Vision. pp. 12849--12858 (2021)

\bibitem{loshchilov2018decoupled}
Loshchilov, I., Hutter, F.: Decoupled weight decay regularization. In:
  International Conference on Learning Representations (2018)

\bibitem{mo2021where2act}
Mo, K., Guibas, L.J., Mukadam, M., Gupta, A., Tulsiani, S.: Where2act: From
  pixels to actions for articulated 3d objects. In: Proceedings of the IEEE/CVF
  International Conference on Computer Vision. pp. 6813--6823 (2021)

\bibitem{mu2021maniskill}
Mu, T., Ling, Z., Xiang, F., Yang, D., Li, X., Tao, S., Huang, Z., Jia, Z., Su,
  H.: {M}ani{S}kill: {G}eneralizable {M}anipulation {S}kill {B}enchmark with
  {L}arge-{S}cale {D}emonstrations. In: Annual Conference on Neural Information
  Processing Systems (NeurIPS) (2021)

\bibitem{park2020non}
Park, J., Joo, K., Hu, Z., Liu, C.K., So~Kweon, I.: Non-local spatial
  propagation network for depth completion. In: European Conference on Computer
  Vision. pp. 120--136. Springer (2020)

\bibitem{peng2018sim}
Peng, X.B., Andrychowicz, M., Zaremba, W., Abbeel, P.: Sim-to-real transfer of
  robotic control with dynamics randomization. In: 2018 IEEE international
  conference on robotics and automation (ICRA). pp. 3803--3810. IEEE (2018)

\bibitem{planche2021physics}
Planche, B., Singh, R.V.: Physics-based differentiable depth sensor simulation.
  In: Proceedings of the IEEE/CVF International Conference on Computer Vision.
  pp. 14387--14397 (2021)

\bibitem{planche2017depthsynth}
Planche, B., Wu, Z., Ma, K., Sun, S., Kluckner, S., Lehmann, O., Chen, T.,
  Hutter, A., Zakharov, S., Kosch, H., et~al.: Depthsynth: Real-time realistic
  synthetic data generation from cad models for 2.5 d recognition. In: 2017
  International Conference on 3D Vision (3DV). pp. 1--10. IEEE (2017)

\bibitem{prakash2019structured}
Prakash, A., Boochoon, S., Brophy, M., Acuna, D., Cameracci, E., State, G.,
  Shapira, O., Birchfield, S.: Structured domain randomization: Bridging the
  reality gap by context-aware synthetic data. In: 2019 International
  Conference on Robotics and Automation (ICRA). pp. 7249--7255. IEEE (2019)

\bibitem{qi2017pointnet++}
Qi, C.R., Yi, L., Su, H., Guibas, L.J.: Pointnet++: Deep hierarchical feature
  learning on point sets in a metric space. Advances in neural information
  processing systems  \textbf{30} (2017)

\bibitem{qu2021bayesian}
Qu, C., Liu, W., Taylor, C.J.: Bayesian deep basis fitting for depth completion
  with uncertainty. In: Proceedings of the IEEE/CVF International Conference on
  Computer Vision. pp. 16147--16157 (2021)

\bibitem{sajjan2020clear}
Sajjan, S., Moore, M., Pan, M., Nagaraja, G., Lee, J., Zeng, A., Song, S.:
  Clear grasp: 3d shape estimation of transparent objects for manipulation. In:
  2020 IEEE International Conference on Robotics and Automation (ICRA). pp.
  3634--3642. IEEE (2020)

\bibitem{schoenberger2016sfm}
Sch\"{o}nberger, J.L., Frahm, J.M.: {Structure-from-Motion Revisited}. In:
  Conference on Computer Vision and Pattern Recognition (CVPR) (2016)

\bibitem{tobin2017domain}
Tobin, J., Fong, R., Ray, A., Schneider, J., Zaremba, W., Abbeel, P.: Domain
  randomization for transferring deep neural networks from simulation to the
  real world. In: 2017 IEEE/RSJ international conference on intelligent robots
  and systems (IROS). pp. 23--30. IEEE (2017)

\bibitem{tremblay2018training}
Tremblay, J., Prakash, A., Acuna, D., Brophy, M., Jampani, V., Anil, C., To,
  T., Cameracci, E., Boochoon, S., Birchfield, S.: Training deep networks with
  synthetic data: Bridging the reality gap by domain randomization. In:
  Proceedings of the IEEE conference on computer vision and pattern recognition
  workshops. pp. 969--977 (2018)

\bibitem{uhrig2017sparsity}
Uhrig, J., Schneider, N., Schneider, L., Franke, U., Brox, T., Geiger, A.:
  Sparsity invariant cnns. In: 2017 international conference on 3D Vision
  (3DV). pp. 11--20. IEEE (2017)

\bibitem{van2019sparse}
Van~Gansbeke, W., Neven, D., De~Brabandere, B., Van~Gool, L.: Sparse and noisy
  lidar completion with rgb guidance and uncertainty. In: 2019 16th
  international conference on machine vision applications (MVA). pp.~1--6. IEEE
  (2019)

\bibitem{wang2019normalized}
Wang, H., Sridhar, S., Huang, J., Valentin, J., Song, S., Guibas, L.J.:
  Normalized object coordinate space for category-level 6d object pose and size
  estimation. In: Proceedings of the IEEE/CVF Conference on Computer Vision and
  Pattern Recognition. pp. 2642--2651 (2019)

\bibitem{weng2021captra}
Weng, Y., Wang, H., Zhou, Q., Qin, Y., Duan, Y., Fan, Q., Chen, B., Su, H.,
  Guibas, L.J.: Captra: Category-level pose tracking for rigid and articulated
  objects from point clouds. In: Proceedings of the IEEE/CVF International
  Conference on Computer Vision. pp. 13209--13218 (2021)

\bibitem{xiong2020sparse}
Xiong, X., Xiong, H., Xian, K., Zhao, C., Cao, Z., Li, X.: Sparse-to-dense
  depth completion revisited: Sampling strategy and graph construction. In:
  European Conference on Computer Vision. pp. 682--699. Springer (2020)

\bibitem{xu2021seeing}
Xu, H., Wang, Y.R., Eppel, S., Aspuru-Guzik, A., Shkurti, F., Garg, A.: Seeing
  glass: Joint point-cloud and depth completion for transparent objects. In:
  5th Annual Conference on Robot Learning (2021)

\bibitem{yue2019domain}
Yue, X., Zhang, Y., Zhao, S., Sangiovanni-Vincentelli, A., Keutzer, K., Gong,
  B.: Domain randomization and pyramid consistency: Simulation-to-real
  generalization without accessing target domain data. In: Proceedings of the
  IEEE/CVF International Conference on Computer Vision. pp. 2100--2110 (2019)

\bibitem{zakharov2019deceptionnet}
Zakharov, S., Kehl, W., Ilic, S.: Deceptionnet: Network-driven domain
  randomization. In: Proceedings of the IEEE/CVF International Conference on
  Computer Vision. pp. 532--541 (2019)

\bibitem{zhang2022close}
Zhang, X., Chen, R., Xiang, F., Qin, Y., Gu, J., Ling, Z., Liu, M., Zeng, P.,
  Han, S., Huang, Z., et~al.: Close the visual domain gap by physics-grounded
  active stereovision depth sensor simulation. arXiv preprint arXiv:2201.11924
  (2022)

\bibitem{zhu2021rgb}
Zhu, L., Mousavian, A., Xiang, Y., Mazhar, H., van Eenbergen, J., Debnath, S.,
  Fox, D.: Rgb-d local implicit function for depth completion of transparent
  objects. In: Proceedings of the IEEE/CVF Conference on Computer Vision and
  Pattern Recognition. pp. 4649--4658 (2021)

\end{thebibliography}
